\documentclass[lettersize,journal]{IEEEtran}
\usepackage{amsmath,amsfonts}
\usepackage{algorithmic}
\usepackage{algorithm}
\usepackage{array}
\usepackage[caption=false,font=normalsize,labelfont=sf,textfont=sf]{subfig}
\usepackage{textcomp}
\usepackage{stfloats}
\usepackage{url}
\usepackage{verbatim}
\usepackage{subcaption}

\usepackage{graphicx}
\usepackage{cite}
\usepackage{xcolor}
\usepackage{booktabs}
\usepackage{multirow}
\definecolor{mygreen}{rgb}{0, 0.5, 0}
\definecolor{myred}{rgb}{0.5, 0, 0}

\title{MPAT: Building Robust Deep Neural Networks against Textual Adversarial Attacks}

\begin{document}

\title{MPAT: Building Robust Deep Neural Networks against Textual Adversarial Attacks}

\author{%
\IEEEauthorblockN{Fangyuan Zhang, Huichi Zhou, Shuangjiao Li, Hongtao Wang\IEEEauthorrefmark{1}}
\IEEEauthorblockA{% 
\\
\textit{School of Control and Computer Engineering, North China Electric Power University, China}\\
\{fyz, zhouhuichi, sjli, wanght\}@ncepu.edu.cn}
\thanks{\IEEEauthorrefmark{1}Hongtao Wang is the corresponding author.}
}

%\markboth{IEEE TRANSACTIONS ON NEURAL NETWORKS AND LEARNING SYSTEMS}%


\maketitle

\begin{abstract}
% 实验主体思想
Deep neural networks have been proven to be vulnerable to adversarial examples and various methods have been proposed to defend against adversarial attacks for natural language processing tasks. However, previous defense methods have limitations in maintaining effective defense while ensuring the performance of the original task. In this paper, we propose a malicious perturbation based adversarial training method (MPAT) for building robust deep neural networks against textual adversarial attacks. 
Specifically, we construct a multi-level malicious example generation strategy to generate adversarial examples with malicious perturbations, which are used instead of original inputs for model training. Additionally, we employ a novel training objective function to ensure achieving the defense goal without compromising the performance on the original task.
We conduct comprehensive experiments to evaluate our defense method by attacking five victim models on three benchmark datasets. The result demonstrates that our method is more effective against malicious adversarial attacks compared with previous defense methods while maintaining or further improving the performance on the original task.
%We conducted comprehensive experiments to evaluate the effectiveness of our defense method by attacking Word Convolutional Neural Network (Word-CNN), Long Short-Term Memory (LSTM), Bidirectional Long Short-Term Memory (BiLSTM), and Bidirectional Encoder Representations from Transformers (BERT) models on three benchmark datasets. The results demonstrate that our method is more effective against malicious adversarial attacks while maintaining or even improving the performance on the original tasks compared to previous Adversarial Training (AT) methods.
% 我们方法的效果
\end{abstract}


\section{Introduction}
%交代研究背景 深度学PAT习框架容易收到攻击
\IEEEPARstart{D}{espite} deep neural networks (DNNs) have shown their prominent performance in natural language processing (NLP) tasks, e.g., text classification, natural language inference, they have been proved to be highly vulnerable to adversarial examples, crafted by adding small and imperceptible perturbations to the input \cite{DBLP:journals/corr/SzegedyZSBEGF13,DBLP:journals/corr/GoodfellowSS14,DBLP:conf/eurosp/PapernotMJFCS16}. In the real world, due to the limited data, the decision boundaries constructed by the model may not fully cover all examples that lie on the same semantic manifold as the training examples (i.e., semantically similar to the training examples). This means that the construction of decision boundaries may be too tight and fail to fully consider potential adversarial examples, thus making the wrong prediction in the face of these examples.
\par Recently, many defense methods have been proposed to defend against adversarial attacks. Most of the work is spent training a DNN model that is equally efficient and reliable in both the original and adversarial examples. In other words, the researchers expect to effectively expand the decision boundaries so that DNN can take into account more potential adversarial examples belonging to the same class as the original example. For different defense strategies, this paper divides the research of adversarial defense into three types. 
\par The first type is \textbf{data augmentation (DA)}, where most methods usually perform a series of transformations on the original examples and augment them into the original training set to increase the diversity of the training data \cite{DBLP:journals/corr/abs-2007-06949,DBLP:conf/emnlp/WeiZ19,dai-adel-2020-analysis,DBLP:conf/iclr/YuDLZ00L18,DBLP:conf/interspeech/PengZZG21,DBLP:conf/iscide/YanLZC19,DBLP:conf/semeval/Daval-FrerotW20,DBLP:journals/corr/abs-2009-10778,DBLP:conf/coling/ZuoCLZ20,DBLP:conf/emnlp/LowellHLW21,DBLP:journals/corr/abs-2007-00875,DBLP:conf/fqas/NugentSL21,DBLP:conf/semweb/PerevalovB20,DBLP:conf/eacl/KoberWBW21}. 
The second type of research focuses on \textbf{representation learning (RL)}, which optimizes and designs embedding models by modifying the original input representation\cite{DBLP:conf/icwsm/LiSLN17,DBLP:conf/ksem/WangW20,DBLP:journals/corr/abs-2006-11627,DBLP:conf/iclr/BelinkovB18,DBLP:journals/corr/abs-1909-06723,DBLP:conf/acl/Malykh19,DBLP:conf/acl/JonesJRL20} or replacing\cite{DBLP:conf/iclr/WangWCGJLL21,DBLP:journals/corr/abs-2009-09587,DBLP:journals/corr/abs-1909-00102,DBLP:journals/corr/abs-2010-12510} the original input representation with representations of semantically similar examples. 
Although both of the above two types of methods demonstrate strong capabilities in defending against adversarial attacks, they also have their respective limitations. In DA methods, the significant increase in training examples will lead to longer training time and higher computational resource requirements. On the other hand, RL methods may excessively emphasize semantic similarity, resulting in the neglect of examples from different categories. In other words, they could cause an excessive expansion of the decision boundary and ultimately decrease the model's performance on the original task.
\par In order to guarantee the performance of the original task, the third type of research, known as \textbf{adversarial training (AT)}, introduces adversarial examples into the training process by adding small perturbations\cite{DBLP:conf/naacl/WangB18,DBLP:conf/ndss/LiJDLW19,DBLP:conf/acl/ZangQYLZLS20,DBLP:conf/acl/HovyKSK18,DBLP:conf/emnlp/XuZYZLS19,DBLP:conf/aaai/LiuZWLC20,DBLP:conf/aaai/LiuLYLSLS20,
%(2) virtual adversarial training
DBLP:conf/acl/ChengJM19,FREEAT,DBLP:conf/iclr/ZhuCGSGL20,DBLP:conf/naacl/PereiraLCPGK21}. Different from traditional DA methods, adversarial training focuses on combining adversarial perturbations with original examples during the training stage, thereby prompting the model to learn more robust representations. 
While adversarial training can guarantee model performance on the original task, it is not well resistant to adversarial attacks\cite{DBLP:conf/emnlp/LiXZLZZCH21}. 
This is because AT tends to constrain perturbations within a small Euclidean ball, typically only able to simulate natural perturbations (which we define as \textbf{benign perturbations}), and fails to effectively mimic the perturbations intentionally added by attackers (defined as \textbf{malicious perturbations}). Therefore, the generated benign adversarial examples cannot effectively expand the decision boundary, thus limiting the effectiveness of adversarial training in resisting adversarial attacks.
\par To overcome the limitations of AT in defending against adversarial attacks, 
%this paper aims to design an adversarial training method that considers both benign and malicious perturbations, ensuring effective defense while maintaining the performance on the original task. 
malicious perturbations should be fully considered as well as benign perturbations in the training of AT methods.
However, it is non-trivial to achieve the objective for the following challenges:
% 多样性 
% 1) The diversity of malicious perturbations. it is hard to construct malicious perturbations to generate malicious adversarial examples that well align with potential adversarial examples;
1) the diversity of malicious perturbations that effectively simulate different variations of potential adversarial examples while maintaining semantic similarity with the original examples; and 
%2) compared with previous adversarial training, combining both malicious and benign perturbations is more challenging since the need to prevent overfitting to the two perturbations and maintain model performance on the original task.
2) the inconsistency between the optimization of benign perturbations within a ball of the Euclidean distance and the addition of malicious perturbations on the semantic manifold space.
%To address the aforementioned issues, this paper proposes a \emph{\textbf{M}alicious \textbf{P}erturbation based \textbf{A}dversarial \textbf{T}raining} method (MPAT), which aims to enhance model robustness and effectively defend against adversarial attacks. 
\par In this paper, we address these challenges by proposing a \emph{\textbf{M}alicious \textbf{P}erturbation based \textbf{A}dversarial \textbf{T}raining} method (MPAT) that aims to ensure effective defense while maintaining performance on the original task. 
Firstly, for the generation of malicious perturbations, we construct a multi-level malicious example generation strategy at the word and sentence levels by designing text paraphrase and synonym replacement rules, and generate a set of malicious adversarial examples. 
%These generated examples lie on the same semantic manifold with the original examples, indicating a good alignment with potential adversarial examples. 
Differing from traditional adversarial training, we utilize the generated malicious adversarial examples instead of the original ones as input and continue to introduce benign perturbations at the embedding layer. 
%Secondly, to effectively combine malicious and benign perturbations without causing the model to overfit to these perturbations, we design an objective function that includes an adversarial loss item which is used to guide the model to learn more robust representations on adversarial examples, and an additional manifold loss item which is introduced to encourage the generated adversarial examples to be on the same semantic manifold as the original examples as much as possible.
Secondly, we design a novel unified objective function for training more robust DNNs while incorporating malicious and benign perturbations. Specifically, this objective function includes an adversarial loss item which is used to guide the model to learn more robust representations on the perturbed adversarial examples, and an additional manifold loss item which is introduced to encourage the perturbed adversarial examples to be on the same semantic manifold as the original examples as much as possible.
%In this way, we balance the defense performance with the original task performance of the model.
\par We extensively experiment on three benchmark datasets for text classification and natural language inference tasks, attacking five victim DNN models (LSTM, Bi-LSTM, Word-CNN, ESIM, and BERT) to evaluate the efficacy of our defense model.
The results show that the proposed MPAT can improve the robustness of the victim model against malicious adversarial attacks, while preserving excellent performance on the original task. 
\par Our contributions are summarized as follows: 
\begin{itemize}
    \item 
   We propose a novel adversarial training method named as MPAT, which effectively combines malicious and benign perturbations and implement adversarial training to build robust model against textual adversarial attacks.
  \item  
   %We construct a multi-level malicious example generation strategy and devise a dynamic algorithm to generate adversarial examples during model training.
   We construct a multi-level malicious example generation strategy and design an additional manifold loss term, such that perturbed examples are distributed on the same semantic manifold as the original input as much as possible.
  \item 
  Extensive experiments are conducted to demonstrate that our MPAT is more effective against malicious adversarial attack compared with previous defense methods, and maintains or further improves the performance on the original task.
\end{itemize}
%剩余工作的介绍
\par The rest of the paper is organized as follows. Section II presents related work, which includes a brief literature review on the field of adversarial studies. The proposed MPAT method is detailed in Section III. Section IV describes the experimental results, comparisons across a range of previous mainstream defense methods, and presents the results of the ablation study. Section V discusses the case study and future work. Conclusions are given in Section VI.
\section{Related Work}
\begin{figure}[t]
\includegraphics[width=0.45\textwidth]{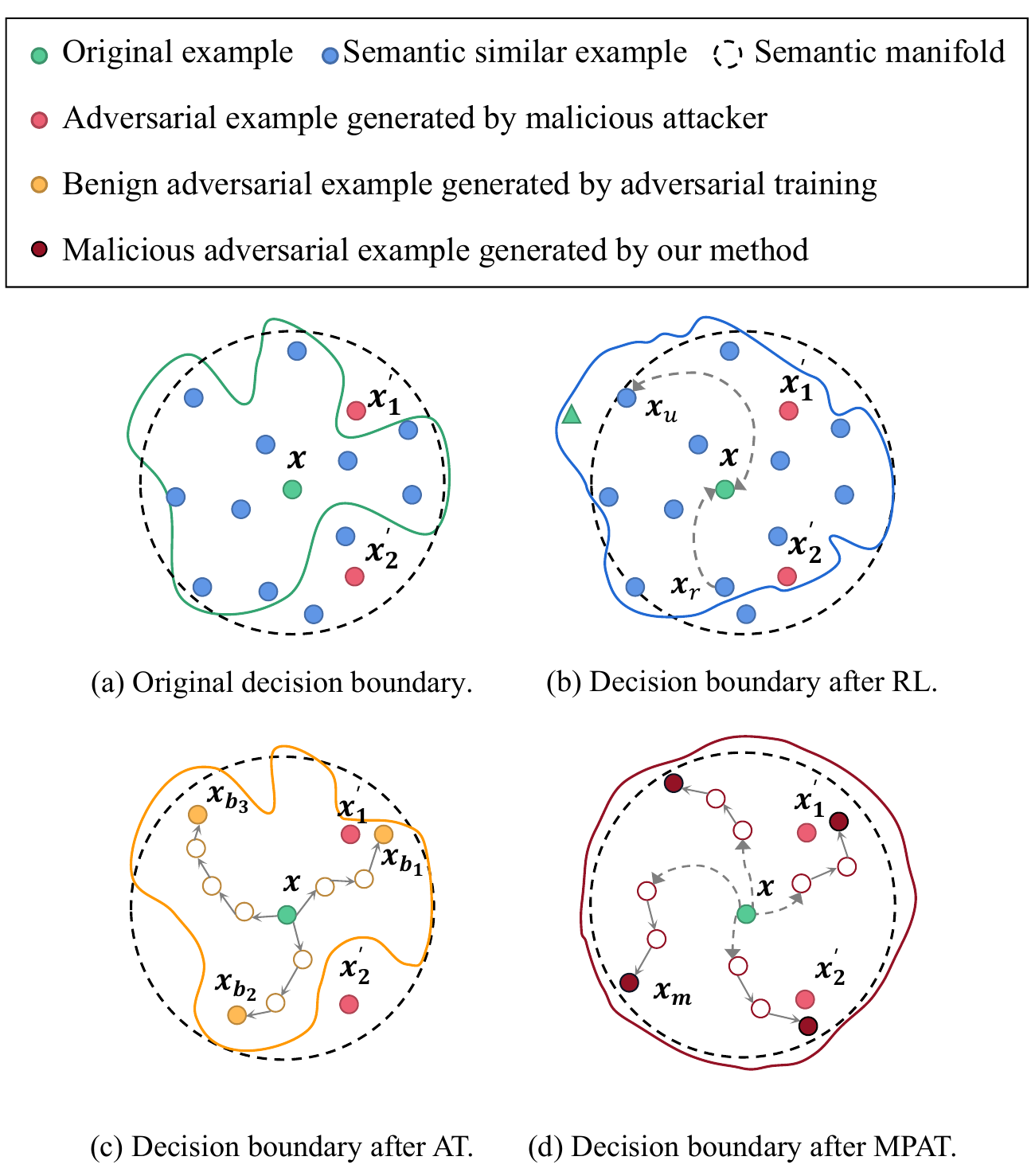}
    %\caption{Decision boundary around original example x. \textcolor{blue}{The neighborhood example (blue circle)} represents an example that is semantically similar to \textcolor{mygreen}{x (green circle)}. For all examples within the decision boundary(irregular border), model can make the same decision as the original example x, that is, making the correct decision. Therefore, the more neighborhood examples (possibly be adversarial) that the decision boundary contains, the better the defense method performance.}
    \caption{The expansion of the model's decision boundary. For convenience, we set the dashed boundary as the semantic manifold where the \textcolor{mygreen}{original example $x$} lies, and the \textcolor{blue}{semantic similar examples} inside the boundary represent examples that are similar to {$x$} in semantics. (a) Ideally, the decision boundary should be smooth and expansive. However, decision boundaries constructed by DNNs are usually wiggly and highly sensitive to adversarial perturbations. (b) The decision boundary is excessively expanded, including examples from other classes. (c) Slight extension of the decision boundary. (d) Effective extension of the decision boundary.}
    \label{fig:Decision_Boundary}
\end{figure}
%\textbf{Data augmentation (DA)} involves expanding the training data to enhance diversity, thereby mitigating the issue of over-fitting. Various methods exist for creating additional data, including rule-based techniques [], example interpolation techniques [], and model-based techniques []. However, generating large and non-iid (non-independent and identically distributed) data is challenging, and the associated computations can be computationally expensive. This is particularly evident in NLP tasks, where ensuring both semantic similarity and grammatical correctness of the augmented data is a complex task.
\subsection{Adversarial Attack}
%It has been widely known that DNNs are highly vulnerable to adversarial perturbations\cite{DBLP:journals/corr/SzegedyZSBEGF13,DBLP:journals/corr/GoodfellowSS14,DBLP:conf/eurosp/PapernotMJFCS16}. By adding small and imperceptible perturbations into input data, these adversarial examples can mislead the predictions of DNNs, leading to wrong classification results or outputs\cite{DBLP:conf/emnlp/ChenSW21,DBLP:conf/eacl/GainskiB23,TextFooler,PWWS,BertAttack}. By adding tiny and imperceptible perturbations to the input data, these adversarial examples can mislead the predictions of DNNs, leading to wrong classification results or outputs.
Initially in the computer vision field, adversarial attacks are extensively explored and comprehensively studied\cite{DBLP:journals/corr/Moosavi-Dezfooli16,DBLP:conf/sp/Carlini017,DBLP:conf/iclr/MadryMSTV18,DBLP:conf/iccv/0009JLZ0WH19}. In the following years, researchers began to pay attention to textual adversarial attack in NLP\cite{PWWS,TextFooler,BertAttack,DBLP:conf/aaai/MaheshwaryMP21,DBLP:conf/naacl/LiZPCBSD21}. 
Suppose there is a DNN model $f(x;\theta)$ with parameter $\theta$  maps an input text $x \in \mathcal{X}$ to a label $y \in \mathcal{Y}$, where $\mathcal{X}$ represents the input text space, and $\mathcal{Y}$ represents the label space. Given an original input $x$, adversarial examples %$\mathcal{X}^{*}$ 
${x}^{*}$ are obtained by designing constrained optimization problems:
\begin{equation}
\label{eq:1}
\mathop{\mathrm{max}}\limits_{{x}^{*}}Sim({x},{x}^{*})~~\textit{s.t.}~~ C(f({x}^{*};\theta))=1
\end{equation}
where $Sim$ is a similarity function between ${x}$ and ${x}^{*}$. $C$ is an adversarial criterion, which equals 1 if ${x}^{*}$ lies outside the decision boundary; otherwise, it equals 0.
As shown in Fig. \ref{fig:Decision_Boundary}(a), the red circles are well-designed adversarial examples, which are semantically similar to the original examples (distributed on the same semantic manifold).  However, due to their positioning outside the decision boundary, they lead the model to make incorrect classifications.
\subsection{Adversarial Defense}
To defend against adversarial attacks, various defense methods have been promoted, including data augmentation, representation learning methods, and adversarial training.

\subsubsection{Data augmentation} DA involves transforming the original text to generate additional synthetic examples, aiming to increase the richness of the original text in scenarios where data is limited. Some works add faint noise to the data without significantly affecting its semantics, such as swapping\cite{DBLP:conf/emnlp/WeiZ19,dai-adel-2020-analysis}, deletion\cite{DBLP:conf/iclr/YuDLZ00L18,DBLP:conf/interspeech/PengZZG21}, or insertion\cite{DBLP:conf/iscide/YanLZC19}. \cite{DBLP:conf/semeval/Daval-FrerotW20,DBLP:journals/corr/abs-2009-10778,DBLP:conf/coling/ZuoCLZ20} replace words in the original text with their synonyms and hypernyms, and \cite{DBLP:conf/emnlp/LowellHLW21,DBLP:journals/corr/abs-2007-00875,DBLP:conf/fqas/NugentSL21} generate new variations by translating the text into another language and then translating it back. Additionally, other works grasp the data distribution and sample new data within it\cite{DBLP:conf/semweb/PerevalovB20,DBLP:conf/eacl/KoberWBW21}. However, DA inevitably increases computational costs such as processing time, storage requirements, and training time.
\subsubsection{Representation learning} RL improves the input representation ability of NLP models and aims to correct the model's representation of adversarial examples. Some works randomly replace the input when training model\cite{DBLP:conf/icwsm/LiSLN17,DBLP:conf/ksem/WangW20,DBLP:journals/corr/abs-2006-11627}, expecting to transform potential adversarial examples “move" towards normal examples that model have seen. Other works force all words and their neighbors to have the same representation in the embedding space by uniformly encoding the input\cite{DBLP:conf/iclr/BelinkovB18,DBLP:journals/corr/abs-1909-06723,DBLP:conf/acl/Malykh19,DBLP:conf/acl/JonesJRL20}. Fig. \ref{fig:Decision_Boundary}(b) shows the expansion of the decision boundary (blue border) by uniformly encoding the original example $x$ and the semantic similar example $x_u$ (the bidirectional arrow process), or replacing the representation of $x$ with the representation of $x_r$ (the unidirectional arrow process). However, RL can lead to an “over-defensiveness" issue, where the model's decision boundary is excessively expanded to include data from another class (e.g., green triangle).
\subsubsection{Adversarial training} The principle of AT is to minimize the model's loss function to optimize its parameters while simultaneously maximizing the difference between adversarial examples and original examples, thereby increasing the model's robustness. AT can be formulated as follows:
\begin{align}
\label{equ:at}
\mathop{\mathrm{argmin}}\limits_{\theta}L_{model}(f(x+\delta^{*};\theta),y),\\
\delta^{*} = \mathop{\mathrm{argmax}}\limits_{\parallel\delta\parallel\leq\epsilon}L_{adv}(f(x+\delta;\theta),y),
\end{align}
where $x$ is the input, $y$ is the true label, $\delta$ and $\delta^{*}$ are adversarial perturbations constrained by a $\epsilon$-ball with radius $\epsilon$, $||\cdot||$ is $L_2$ norm, $L_{adv}$ is the loss for generating adversarial perturbation, and $L_{model}$ is the loss for optimizing the parameters of the model. 
\par Following Equation (\ref{equ:at}), many variants of the AT method are proposed by designing $L_{model}$ and $L_{adv}$. \cite{DBLP:conf/naacl/WangB18,DBLP:conf/ndss/LiJDLW19,DBLP:conf/acl/ZangQYLZLS20} attempt to directly use the generated adversarial examples and the original training examples together to train a robust DNN model that can resist attacks. Inspired by gradient-based adversarial attacks in the computer vision field, other researchers dynamically generate adversarial examples in the input space\cite{DBLP:conf/acl/HovyKSK18,DBLP:conf/emnlp/XuZYZLS19,DBLP:conf/aaai/LiuZWLC20,DBLP:conf/aaai/LiuLYLSLS20}, or in the embedding space\cite{DBLP:conf/acl/ChengJM19,FREEAT,DBLP:conf/iclr/ZhuCGSGL20,DBLP:conf/naacl/PereiraLCPGK21}.
Although AT can ensure or even improve the accuracy of the original task, it cannot well resist malicious adversarial attack \cite{DBLP:conf/emnlp/LiXZLZZCH21}. Specifically shown in Fig. \ref{fig:Decision_Boundary}(c), there is a benign adversarial example $x_{b_{1}}$ (yellow circle) near $x_{1}'$. After the model is trained on $x_{b_{1}}$, the decision boundary (yellow border) can be expanded to include $x_{1}'$, that is, the defense is successful. But for $x_{2}'$, adversarial training does not generate benign adversarial examples in its vicinity and fails to effectively expand the decision boundary. Therefore, AT fails to defend $x_{2}'$. Note that we refers to adversarial training as \textbf{Benign Perturbation based Adversarial Training (BPAT)}.
\begin{figure*}[t]
\includegraphics[width=\textwidth]{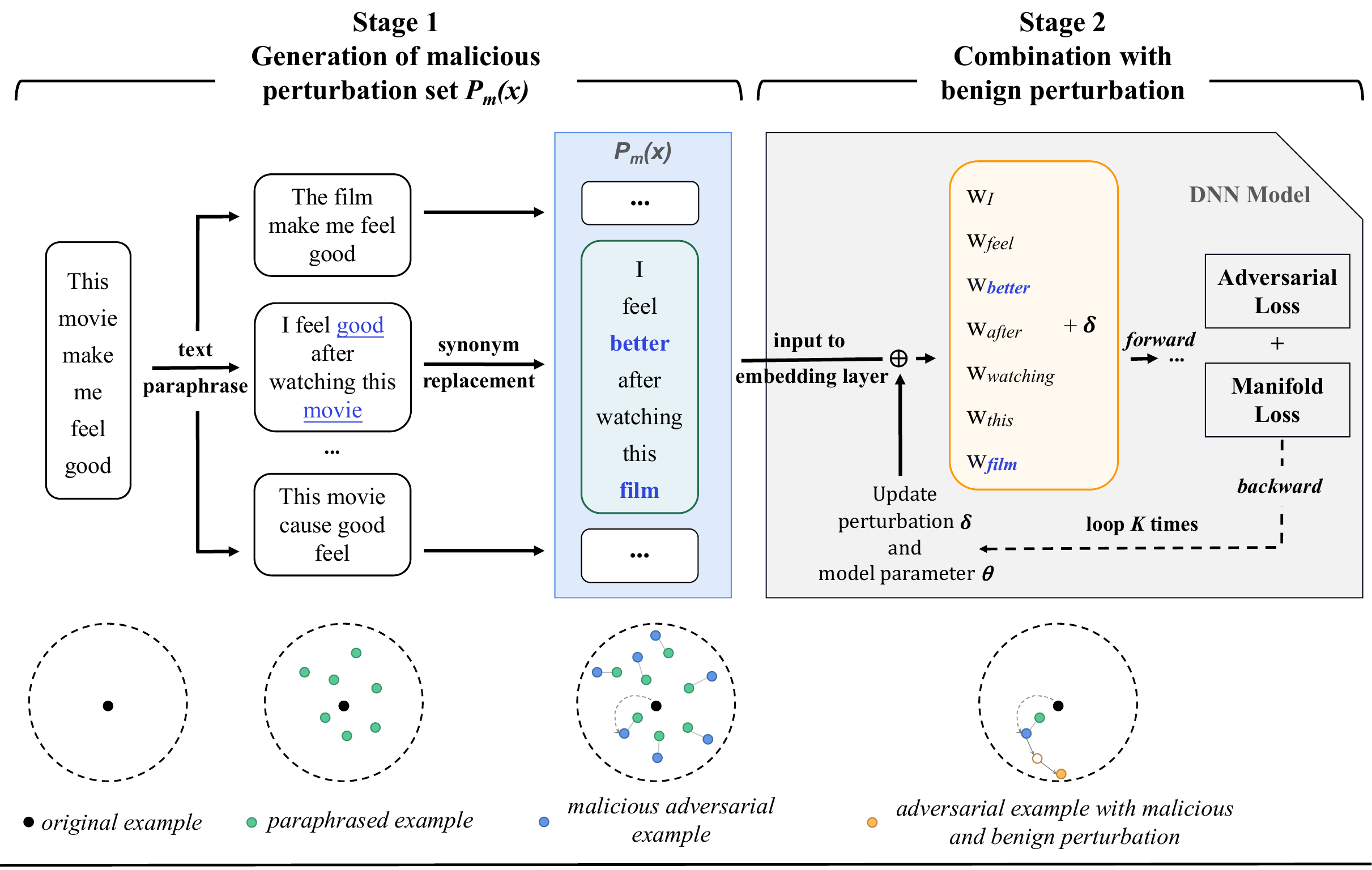}
    \caption{An illustration of our MPAT in training process. 
    $[W_{I},W_{feel},...,W_{film}]$ represents the embedding output of the input sequence. The upper part describes the workflow of each training epoch, and the lower part shows the examples generated corresponding to each step.}
    \label{fig:Framework}
\end{figure*}
\section{Method}
%In this section, we first introduce the process of the previous adversarial training, and then we illustrate our MPAT defense method in detail. Finally, we describe the overall training process of the model.
In this section, we first introduce the motivation for the proposed MPAT defense method, then mathematically define the task, and finally we detail our MPAT and describe the overall training process of the model.
\subsection{Motivation}
Different from the previous defense methods, our MPAT aims to effectively defend against malicious adversarial attacks while ensuring the performance on the original task. In other words, MPAT generates malicious adversarial examples that are distributed as much as possible on the same semantic manifold as the original examples in the feature representation space. By such generation, we effectively expand the model's decision boundary to include more potentially adversarial examples belonging to the same class as the original examples. 
As shown in Fig. \ref{fig:Decision_Boundary}(d), the generated malicious adversarial examples are highly likely to be potential adversarial examples as they are located near the decision boundary. As a result, training model on $x_m$ can effectively expand the decision boundary around (crimson border), and model can make correct decision for the adversarial examples $x_{1}'$ and $x_{2}'$ inside the decision boundary.
\subsection{Task Definition}
Suppose there is a DNN model $f(x;\theta)$ maps an input text space $\mathcal{X}$ to a label space $\mathcal{Y}$, and the model parameter is $\theta$. Let $x_n \in \mathcal{X}$ be the input text and $y_n \in \mathcal{Y}$ be its corresponding true label. In text classification, $x_n$ denotes a single sentence, while in natural language inference, $x_n$ consists of a text pair with two sentences, denoted as $x_n=(premise, hypothesis)$. The main goal of normal model training is to minimize the loss function $L_{model}$ by adjusting the model parameter $\theta$ so that the predicted result is as close as possible to the true label $y_n$. In the case of $N$ data points, the training goal is to solve:
%\begin{equation}
%\mathop{min}\limits_{\theta}\mathbb{E}_{(x,y)\sim D}\left[ L(f(x;\theta),y) \right]
%\end{equation}
\begin{equation}
\mathop{\mathrm{min}}\limits_{\theta}\frac{1}{{N}}\sum\limits_{n}L_{model}(f(x_n;\theta),y_n).
\end{equation}
%\begin{equation}
%\mathop{min}\limits_{\theta}\frac{1}{{N}}\sum\limits_{n} \mathop{max}\limits_{\parallel\delta\parallel\leq\epsilon}L(f(x_n+\delta_n;\theta),y_n),
%\end{equation}
\par Most BPAT methods design an adversarial loss function $L_{adv}$ to optimize the parameter $\theta$ and minimize the maximum risk when adding benign perturbation to the original input. BPAT can be formulated as follows:
%\begin{equation}
%\mathop{\mathrm{min}}\limits_{\theta}\frac{1}{{N}}\sum\limits_{n} \mathop{\mathrm{max}}\limits_{\parallel\delta_n\parallel\leq\epsilon}L_{model}(f(x_n+\delta_n;\theta),y_n),
%\end{equation}
\begin{align}
\mathop{\mathrm{min}}\limits_{\theta}L_{adv}(\theta)&=\mathop{\mathrm{min}}\limits_{\theta}L_{adv}(f(x_n+\delta_{n};\theta),y_n)\nonumber\\
&=\mathop{\mathrm{min}}\limits_{\theta}\frac{1}{{N}}\sum\limits_{n} \mathop{\mathrm{max}}\limits_{\parallel\delta_n\parallel\leq\epsilon}L_{model}(f(x_n+\delta_n;\theta),y_n),
\end{align}
%\begin{equation}  \mathop{min}\limits_{\theta}\mathbb{E}_{(x,y)\sim D}\left[ \mathop{max}\limits_{\parallel\delta\parallel\leq\epsilon}L(f(x+\delta;\theta),y)\right],
%\end{equation}
%再次指出之前方法的弊端
where $\delta_n$ is the benign perturbation and $x_n$ is the original input. As mentioned above, due to the constraint of the gradient direction-based radius $\epsilon$, excessive perturbation may cause the high-dimensional embeddings to lose their original semantic information, resulting in difficulties in model convergence and lack of generalization ability. Therefore, $\delta_n$ is usually constrained within a small radius for iterative optimization. On the other hand, although small perturbations can preserve semantic similarity, they may not effectively expand the decision boundary for adversarial examples, making it challenging for BPAT to find suitable and stable hyper-parameters $\delta^{*}$ to train DNN models that can effectively defend against adversarial attacks.

\subsection{Framework of MPAT}
%介绍分类器范式
%Suppose there is a text classifier $f(X;\theta):X\rightarrow{Y}$ with parameter $\theta$, where $X$ and $Y$ are denoted as the input and output space respectively. The similar to NLI model $f(X_1, X_2;\theta):[X_1, X_2]\rightarrow{Y}$, where $X_1$, $X_2$ and $Y$ are denoted as the premise, hypothesis and output space respectively. 

%介绍对抗训练的范式
%Given a text-label pair$(x \in X, y \in Y)$ satisfying the distribution $D$ and the loss function $L(\cdot, \cdot)$, \textbf{previous adversarial training} aims to optimize parameter $\theta$ to minimize the maximum risk when adding perturbation $\delta$ to the original input x as:
%提出我们方法的不同
Different from BPAT, our MPAT constructs a multi-level adversarial example generation strategy and collects the generated examples into a malicious perturbation set $P_m(x)$. We implement adversarial training on $x' \in P_m(x)$ instead of the original input $x$.
Meanwhile, to ensure that $x'$ is distributed as closely as possible to the semantic manifold of the original $x$, we introduce an additional manifold loss term in the adversarial loss function. Assuming we have a DNN with $L+1$ layers, where the final layer is the softmax layer, we denote the activation of the $L$-th layer as $a^{(L)}$, which approximately represents the low-dimensional embedding of the text on the manifold. Then, the training process of MPAT on $N$ data points can be abstracted as follows:
%\begin{equation}
%\label{eq:2}
%\mathop{min}\limits_{\theta}\mathbb{E}_{(x,y)\sim D,x_m\in P_m(x)}\left[ \mathop{max}\limits_{\parallel \delta \parallel \leq \epsilon}L(f(x+\delta;\theta),y)\right]
%\end{equation}
\begin{equation}
\label{equ:mpat}
\mathop{\mathrm{min}}\limits_{\theta}\frac{1}{{N}}\sum\limits_{n} \mathop{\mathrm{max}}\limits_{\parallel\delta_n\parallel\leq\epsilon}\{L_{model}(f(x_{n}'+\delta_n;\theta),y_n)+\lambda G(x_n,x_{n}')\},
\end{equation}
\begin{equation}
\label{equ:g}
G(x_n,x_{n}') = \frac{1}{2}||a_n^{(L)}-a_{n}'^{(L)}||_{2}^{2},
\end{equation}
\begin{equation}
\label{equ:xn'}
x_{n}' = {RandomSample}(P_m(x_{n})),
\end{equation}
where $x_n$ and $y_n$ are examples from the original dataset, $x_{n}'$ is a sample from the generated malicious perturbed set, $\delta_n$ represents the perturbation added in the embedding layer, $\epsilon$ is a clipping parameter used to constrain the magnitude of the perturbation, $G$ is a manifold loss and $\lambda$ is a hyper-parameter for the manifold loss, and $a_n^{(L)}$ and $a_{n}'^{(L)}$ represent the activations of examples $x_n$ and $x_{n}'$ at the $L$-th layer, respectively. $RandomSample$ in Equation (\ref{equ:xn'}) represents the random sampling of an adversarial example $x_{n}'$ from the malicious perturbation set $P_m(x)$.
%解释步骤

To realize the objective in Equation (\ref{equ:mpat}), (\ref{equ:g}) and (\ref{equ:xn'}), our method can be divided into two stages.
%第一步
\textbf{In the first stage} (Section \ref{sec:C}), 
we define a series of text paraphrase and synonym replacement rules to generate malicious perturbation set $P_m(x)$ at the sentence-level and word-level, respectively. As illustrated in Stage 1 in Fig. \ref{fig:Framework}, we first paraphrase the original example to generate multiple paraphrased  examples (green circles). Then, we randomly replace some words in these examples with their synonyms. To ensure sentence smoothness, we set a perplexity (PPL) constraint, requiring the generated examples to have a PPL value below a certain threshold. The final generated multiple malicious perturbed examples (blue circle) will be collected into a set $P_m(x)$. 
%we generate malicious perturbation set $P_m(x)$ using text paraphrase and synonym replacement respectively at the sentence-level and word-level. As illustrated in the left half of Fig. \ref{fig:Framework}, the original sample is first paraphrased to generate multiple paraphrased  examples (green circles). Then some words in these examples are randomly replaced with synonyms. To ensure smoothness of the sentence, it is necessary for these generated examples to have a perplexity (PPL) below a certain threshold. The final generated multiple malicious perturbed examples (blue circle) will be collected into a set $P_m(x)$. 

%第二步
\par \textbf{The second stage} (Section \ref{sec:D}) effectively combines malicious and benign perturbations. 
%continues to add benign perturbation $\delta$ to $x' \in P_m(x)$ at the embedding-level.
%, and fine-tune $\delta$ to obtain the optimal perturbation within the $\epsilon$-ball constraint.
%In \textbf{the second stage} (\ref{sec:C}), 
%Subsequently, malicious perturbation based adversarial training is conducted with the objective of finding the optimal model parameters $\theta$ while fine-tuning $\delta$ to minimize the loss function. 
As shown in Stage 2 in Fig. \ref{fig:Framework}, we randomly select one sample $x'$ in the set $P_m(x)$, and input it into the model instead of $x$. Then, we add perturbation $\delta$ to $x'$ in the embedding layer. Based on the generated adversarial examples $x' + \delta$ (yellow circle) combining malicious and benign perturbations, we implement adversarial training and update the $\delta$ iteratively by the gradient of the loss with respect to input $x'$. In this way, the model can see more potential adversarial examples that it has not seen before, and can correctly distinguish their labels, effectively extending the decision boundary. 

    \begin{algorithm*}[t]
        \caption{Generation of malicious perturbation set $P_m(x)$}
        \label{alg:1}
        \begin{algorithmic}[1] %这个1 表示每一行都显示数字
        \REQUIRE  %算法的输入参数：Input
            Input sentence $x$, the replacement percentage $r\%$
            \STATE  Initialize $P_m(x)\leftarrow\{x\}$;
            \STATE  Initialize syntactic labels of single word $L\leftarrow\{ADV,ADJ,N,V,PREP,PRON,ABBR\}$;
            \STATE  Get the constituents $c(i,j,q)$ of $x$ containing more than two words ;
            \FOR{each $c(i,j,q),l\notin L$} 
            \STATE Remove $c(i,j,q)$ to obtain the rest unchanged part of the sentence $x_{rest}$;\\
            \STATE Paraphrase $c(i,j,q)$ using paraphrase model: $c'(i,j,q)\leftarrow Para(c(i,j,q))$;\\
            \STATE Get complete paraphrased example: $x_{r}\leftarrow c'(i,j,q) \oplus x_{rest}$
            \STATE $P_m(x)\leftarrow P_m(x)\cup x_{r}$;\\
            \ENDFOR
            \STATE Generate substitution candidate for each word in $P_m(x)$;
            
            \FOR{$p_{m_{i}}\in P_m(x)$} 
            \IF{$PPL(p_{m_{i}})>PPL(x)$} 
            %\COMMENT{\textcolor{green}{aa}}
            \STATE Remove $p_{m_{i}}$ from $P_m(x)$; 
    ~~~~\textcolor{mygreen}{/*Filter the malicious perturbation set*/}
            \ELSE
            \STATE  Sample $r*|p_{m_{i}}|$ words;~~~~\textcolor{mygreen}{/*Select the word to be replaced*/}
            \FOR{$word$ $\in$ $sampled$ $words$}
            \STATE  Randomly select one synonym in the substitution candidate; 
            \STATE  $word\leftarrow synonym$;
            \ENDFOR
            \ENDIF 
            \ENDFOR
            \RETURN $P_m(x)$; %算法的返回值
        \end{algorithmic}
    \end{algorithm*}

\begin{figure}[t]
    \centering
    \includegraphics[width=0.8\linewidth]{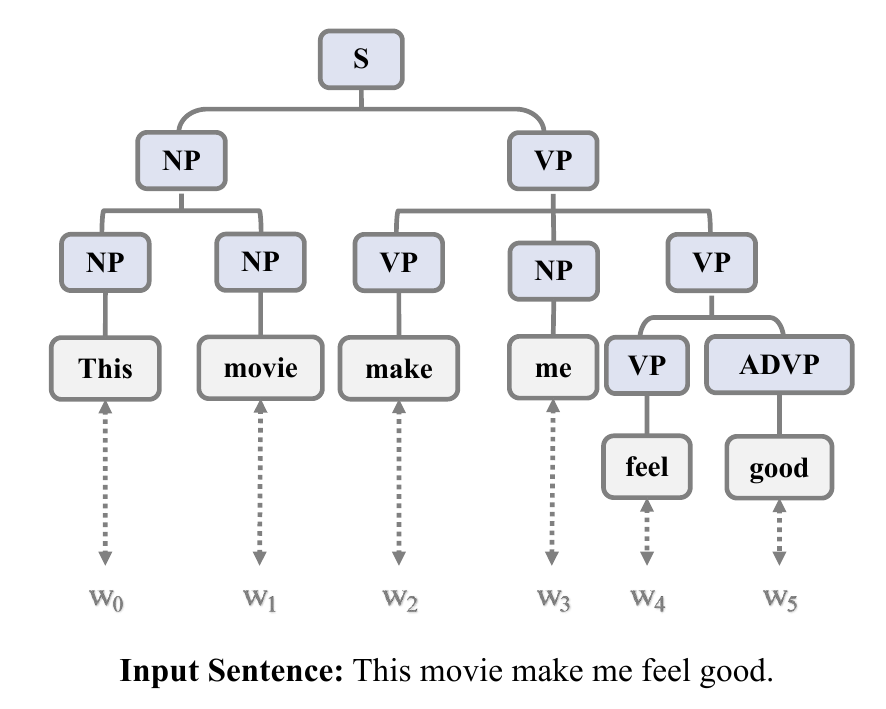}
    \caption{A constituency parsing tree, 
     where the blue box represents
    the syntactic label and the gray box represents the original word.
    Then S, VP, NP and ADVP refer to a sentence, a verb phrase, a noun
    phrase and an adverb phrase respectively. In order to facilitate the
    display, only the syntactic label is shown in the figure, which simplifies the label of single word.}
    \label{fig:constituency parsing tree}
\end{figure}
\subsection{Generation of $P_m(x)$}
\label{sec:C}

Formally, let $x = [w_0, w_1, ..., w_{n-1}]$ be a sentence consisting of $n$ words. To introduce sentence-level perturbations while preserving its semantics, we perform paraphrasing on $x$. To enhance diversity in the generated examples, we not only paraphrase the entire sentence but also each constituent within the sentence. In the following, we provide a detailed description of the steps involved in the generation of $P_m(x)$.
\par\textbf{Step 1: Text paraphrase}. We perform constituency parsing \cite{SuPar}, which can organize $x$ into nested constituents, i.e., constituency parsing tree composed of  phrases. We denote $c(i,j,q)$ as a constituent spanning $w_i...w_j$ with its syntactic label $q$. As show in Fig. \ref{fig:constituency parsing tree}, the constituency parsing tree of the given example is constructed. In this tree, $c(0, 1, NP)$ represents a noun phrase spanning $w_0,w_1$, e.g., “This movie". $c(5, 5, ADVP)$ represents an adverbial phrase consisting of only one word “good". For each $c(i,j,q)$, except single word phrase, we employ style-transfer
based paraphrase model \cite{DBLP:conf/emnlp/KrishnaWI20} to modify text while keeping the original meaning and generate the paraphrased constituent spanning $c'(i,j,q) = Para(c(i,j,q)), q \notin Q$, where $Para$ represents paraphrase model and $Q$ represents syntactic labels of single word, e.g., $N, V, ADB,$ etc. The corresponding paraphrased example $x_r$  is the splicing of $c'(i,j,q)$ and the rest unchanged part $x_{rest}$:
\begin{equation}
    x_r = c'(i,j,q) \oplus x_{rest}
\end{equation}

As shown in Fig. \ref{fig:Framework}, after rewriting “This movie” as “This film”, the sentence becomes “This film make me feel good”. Since splicing will lead to grammatical errors, we filter out the paraphrased  examples with a perplexity (PPL) higher than the threshold (set as the PPL of the original sentence $x$). We collect the paraphrased examples generated in this rule into a set denoted as $P_m(x) = \{x_r^{(1)},x_r^{(2)}, ..., x_r^{(m)}\}$.

\textbf{Step 2: Synonym replacement.} For word-level perturbation,  we perform synonym replacement on each paraphrased  example in $P_m(x)$. Specifically, for $x^{(k)}_r$ , we sample words to be replaced with a certain rate and build suitable substitution candidate that contains all synonyms of these words by WordNet \footnote{WordNet is a large lexical database of English. For more information, please go https://wordnet.princeton.edu/}. Then, we randomly select one synonym to replace original word. In this way, a two-level malicious perturbation set $P_m(x)$ is generated, which contains more potential adversarial examples. And the input to the model to participate in the training process is replaced from the given $x$ to a maliciously perturbed example $x'$ selected from $P_m(x)$. The generation procedure is summarized in Algorithm \ref{alg:1}. Note that we also add the original example $x$ to $P_m(x)$ to guarantee the possible case without any malicious perturbation.

\begin{algorithm*}[t]
    \caption{Malicious Perturbation based Adversarial Training method}
    \label{alg:2}
    \begin{algorithmic}[1] %这个1 表示每一行都显示数字
    \REQUIRE  %算法的输入参数：Input
        Training data $D = \{x_i, y_i\}_{N}$, model parameter $\theta$, ascent step $K$, perturbation bound $\epsilon$, learning rate $\tau$
        \STATE Initialize $\theta$
        \STATE $\delta \leftarrow 0$
        \FOR{$epoch = 1\cdots N_{ep}/K$}
        \FOR{$minibatch$  $B \subset D$}
        \FOR{$original$ $example$ $x \in B$}
        \STATE Get the  malicious perturbation set $P_m(x)$ of $x$ according to Algorithm \ref{alg:1};
        \STATE Randomly select one sentence $x'$ in $P_m(x)$;
        \STATE Replace $x$ with $x'$;
        \ENDFOR
        \FOR{$i=1\cdots K$}
        \STATE Calculate $g_{adv}$ as defined in Equation (\ref{equ:gadv});
        \STATE Use gradients calculated for the minimization step to update $\delta$;
        \STATE ~~~~ $\delta\leftarrow\delta+\epsilon\cdot sign(g_{adv})$
        \STATE ~~~~ $\delta\leftarrow clip(\delta,-\epsilon,\epsilon)$  ~~~~~~~\textcolor{mygreen}{/*Constrain $\delta$ by a $\epsilon$-ball with radius $\epsilon$*/}
        \STATE Update $\theta$ with stochastic gradient descent;
        \STATE ~~~~ %$g_{\theta}\leftarrow\mathbb{E}_{(x,y)\in B}[\nabla_{\theta}L_{model}(f(x'+\delta;\theta),y)+\lambda\nabla_{\theta}G(x,x')]$
        Calculate $g_{\theta}$ as defined in Equation (\ref{equ:gtheta});
        \STATE ~~~~ $\theta\leftarrow \theta-\tau g_{\theta}$ 
        \ENDFOR
        \ENDFOR
        \ENDFOR
    \end{algorithmic}
\end{algorithm*}
\subsection{Combination of benign and malicious perturbations}
\label{sec:D}
The entire training process is illustrated in Algorithm \ref{alg:2}. In every epoch of the training process, we follow the proposed multi-level malicious adversarial example generation strategy and generate a malicious perturbation set $P_m(x)$ for each original input $x$. Then, we randomly select an adversarial example $x'$ from $P_m(x)$ so that the model will be trained on $x'$ instead of $x$ (Lines 6-8). 
%Then, we perform $K$ descent steps on the model parameter $\theta$ together with $K$ iterations ascent steps on the benign perturbation $\delta$ to update $\delta$ and $\theta$ simultaneously (Lines 10-17).
Then, we perform $K$ descent steps on the parameter of the model $\theta$ and $K$ ascent steps on the benign perturbation $\delta$ to update $\delta$ and $\theta$ simultaneously (Lines 10-17).
%First, We use gradients to update $\delta$ and constrain them in a small normalization ball to keep the perturbations minimum. 
Specifically,
%the perturbation $\delta$ is calculated by multiple iterations of gradient ascent, while the parameter $\theta$ is calculated by multiple iterations of gradient descent. 
we take the following step 
%(with the perturbation bound $\epsilon$) 
in $K$ iterations:
\begin{equation}
\delta^{(t+1)}=\delta^{(t)}+\epsilon\cdot sign(g_{adv}),
\label{equ:delta}
\end{equation}
\begin{equation}
\theta^{(t+1)}=\theta^{(t)}-\tau\cdot g_{\theta},
\label{equ:theta}
\end{equation}
where $\epsilon$ is the perturbation bound, $\tau$ is the learning rate, $sign(\cdot)$ is the indicator function. $g_{adv}$ is the gradient of the loss with respect to the input $x$, and $g_{\theta}$ represents the gradient of the loss with respect to the input $\theta$:
%\begin{equation}
%g_{adv} = \nabla_{x}L_{model}(f(W_{x'}+\delta^{(t)},\theta),y)
%\label{equ:gadv}
%\end{equation}
\begin{equation}
g_{adv} = \nabla_{x}L_{model}(f(x'+\delta^{(t)},\theta^{(t)}),y),
\label{equ:gadv}
\end{equation}
\begin{equation}
 g_{\theta}=\nabla_{\theta}L_{model}(f(x'+\delta^{(t)};\theta^{(t)}),y)+\lambda\nabla_{\theta}G(x,x').
\label{equ:gtheta}
\end{equation}
%where $W_{x'}$ is the embedding output of the input sequence $x'$ and $\theta$ is the model parameter. The finally generated adversarial example in the embedding space $W_{x'}+\delta_{K}$ will be used for model training.

In addition, we use Frobenius normalization to constrain the perturbation $\delta$ (Line 14), thereby limiting its magnitude and preventing the model from diverging: 
%\begin{equation}
%\delta^{(t)} = \Pi_{\parallel \delta^{(t)} \parallel\leq\epsilon}, t = 1, ..., K.
%\label{eq6}
%\end{equation}
\begin{equation}
\delta^{(t)} = clip(\delta^{(t)},-\epsilon,\epsilon), t = 1, ..., K,
\label{eq6}
\end{equation}
where $clip$ represents the process that projects $\delta^{(t)}$ in each iteration onto the $\epsilon$-ball. 
%The finally generated adversarial example in the embedding space ${x'}+\delta^{(K)}$ will be used for model training.
%\par In each epoch of training, the parameters $\theta$ of the model are updated through the gradient descent method, and the specific steps are as follows. First, calculate the gradient of the loss function with respect to the model parameters $\theta$:

%\subsection{Overall Training Process}
%Then, using the calculated gradient $g_{\theta}$, the gradient descent method is used to update the model parameters $\theta$:
\section{Experiments}
In this section, we conducted extensive experiments to evaluate the performance of the proposed MPAT method. We start by introducing the experimental setup, followed by presenting the experimental results on two common NLP tasks. All experiments were conducted using two GeForce RTX 2080Ti 11-GB GPUs and an Intel(R) Core(TM) i9-9900K CPU @ 3.60GHz.

\subsection{Experimental Settings}

\subsubsection{Datasets} 
We focus on three prevailing real-world datasets on text classification (TC) task and natural language inference (NLI) task to evaluate defense performance. Dataset details are shown in Table \ref{tab:data}.
\begin{itemize}
      \item 
      \textit{IMDB}\footnote{https://datasets.imdbws.com/}: Movie review dataset for sentiment classification, comprising movie reviews from the Internet Movie Database along with associated sentiment labels.
      \item    \textit{AGNEWS}\footnote{https://huggingface.co/datasets/agnews}: News articles categorized into four different topics: world, sports, business, and science/technology.
      \item    \textit{SNLI}\footnote{https://nlp.stanford.edu/projects/snli/}: A large-scale dataset for natural language inference used to determine whether a premise entails, contradicts, or is independent of a hypothesis. 
\end{itemize}
\begin{table}[t]
    \centering
    \caption{Details and preprocessing settings for each dataset. \#L is the padding length, \#C is the number of classes, Vocab means the size of vocabulary, and Train/Test shows the number of training and testing examples the dataset contains.}
    \begin{tabular}{lccccc}
    \specialrule{.1em}{0em}{.1em} %加粗
        \textbf{Dataset} & \textbf{Task}& \textbf{\#L} & \textbf{\#C} & \textbf{Vocab}& \textbf{Train/Test}\\
        \hline
        IMDB  & TC & 300 & 2 & 80k & 25k/25k\\
        AGNEWS & TC & 50 & 4 & 80k & 120k/7.6k\\
        SNLI & NLI & 15 & 3 & 80k & 550K/10k\\
    \specialrule{.1em}{0em}{.1em}
    \end{tabular}
    \label{tab:data}
\end{table} 
\subsubsection{Victim Models} We use the following five deep neural networks as our victim models. (\romannumeral1) Word Convolutional Neural Network (Word-CNN) with two embedding layers, three convolutional layers with filter sizes of 3, 4, and 5,  one 1D-max-pooling layer, and a fully connected layer. (\romannumeral2) Long Short-Term Memory (LSTM), and (\romannumeral3) Bidirectional Long Short-Term Memory (Bi-LSTM) which consists of a 100-dimensional embedding layer, two LSTM or bidirectional LSTM layers and a fully connected layer. (\romannumeral4) Enhanced Sequential Inference Model (ESIM) which combines attention mechanism and Bi-LSTM under the above configuration. (\romannumeral5) BERT is a powerful language representation model that utilizes a transformer architecture with 12 layers and 768 hidden size.

\begin{table*}[t]
    \caption{Results for the defense performance. The best results are in \textbf{bold}. ACC$_{clean}$, ACC$_{test}$, and ACC$_{adv}$ refer to the victim model accuracy on clean examples, test examples, and adversarial examples, respectively. $*$+MPAT represents the victim model after being defended using our MPAT defense method. $\downarrow$ ($\uparrow$) means that the lower (higher) value is the better.}
    \centering
    \begin{tabular}{llllllllll}
        \specialrule{.1em}{0em}{.1em}
        \multicolumn{2}{l}{\textbf{Attack Method}}  
             & \multicolumn{2}{c}{\textbf{No Attack(\%)}} 
             & \multicolumn{2}{c}{\textbf{PWWS(\%)}} 
             & \multicolumn{2}{c}{\textbf{TextFooler(\%)}} 
             & \multicolumn{2}{c}{\textbf{BertAttack(\%)}}    \\ 
         \hline
         \textbf{Dataset}             
             & \textbf{Victim Model} 
             & \textbf{ACC$_{clean}\uparrow$} 
             & \textbf{ACC$_{test}\uparrow$} 
             & \textbf{ACC$_{adv}\uparrow$} 
             & \textbf{ASR$\downarrow$} 
             & \textbf{ACC$_{adv}\uparrow$} 
             & \textbf{ASR$\downarrow$} 
             & \textbf{ACC$_{adv}\uparrow$} 
             & \textbf{ASR$\downarrow$}  \\ 
        \hline
        \multicolumn{1}{l}{\multirow{8}{*}{{IMDB}}}  
            & \multicolumn{1}{l}{{LSTM}} 
            & \multicolumn{1}{c}{{90.6}} 
            & \multicolumn{1}{c}{{87.5}} 
            & \multicolumn{1}{c}{{21.6}} 
            & \multicolumn{1}{c}{{76.1}} 
            & \multicolumn{1}{c}{{67.6}} 
            & \multicolumn{1}{c}{{15.6}} 
            & \multicolumn{1}{c}{{25.6}} 
            & \multicolumn{1}{c}{{71.7}} \\
        \multicolumn{1}{c}{}  
            & \multicolumn{1}{l}{{LSTM+MPAT}} 
            & \multicolumn{1}{c}{{\textbf{93.1}}} 
            & \multicolumn{1}{c}{{\textbf{87.6}}} 
            & \multicolumn{1}{c}{{\textbf{83.9}}} 
            & \multicolumn{1}{c}{{\textbf{10.3}}} 
            & \multicolumn{1}{c}{{\textbf{88.6}}} 
            & \multicolumn{1}{c}{{\textbf{8.7}}} 
            & \multicolumn{1}{c}{{\textbf{90.6}}} 
            & \multicolumn{1}{c}{{\textbf{11.1}}} \\
        %\cline{3-10}
        \multicolumn{1}{c}{}  
            & \multicolumn{1}{l}{{Bi-LSTM}} 
            & \multicolumn{1}{c}{{\textbf{89.0}}} 
            & \multicolumn{1}{c}{{93.4}} 
            & \multicolumn{1}{c}{{26.5}} 
            & \multicolumn{1}{c}{{71.6}} 
            & \multicolumn{1}{c}{{82.2}} 
            & \multicolumn{1}{c}{{17.8}} 
            & \multicolumn{1}{c}{{29.5}} 
            & \multicolumn{1}{c}{{66.9}} \\
        \multicolumn{1}{c}{}  
            & \multicolumn{1}{l}{Bi-LSTM+MPAT}
            & \multicolumn{1}{c}{{88.2}} 
            & \multicolumn{1}{c}{{\textbf{96.9}}} 
            & \multicolumn{1}{c}{{\textbf{86.9}}} 
            & \multicolumn{1}{c}{{\textbf{8.83}}} 
            & \multicolumn{1}{c}{{\textbf{91.1}}} 
            & \multicolumn{1}{c}{{\textbf{8.8}}} 
            & \multicolumn{1}{c}{{\textbf{93.7}}} 
            & \multicolumn{1}{c}{{\textbf{7.9}}} \\
        %\cline{3-10}
        \multicolumn{1}{c}{}  
            & \multicolumn{1}{l}{{Word-CNN}} 
            & \multicolumn{1}{c}{{83.8}} 
            & \multicolumn{1}{c}{{{91.4}}} 
            & \multicolumn{1}{c}{{24.2}} 
            & \multicolumn{1}{c}{{59.5}} 
            & \multicolumn{1}{c}{{69.1}} 
            & \multicolumn{1}{c}{{33.2}} 
            & \multicolumn{1}{c}{{36.9}} 
            & \multicolumn{1}{c}{{52.6}} \\
        \multicolumn{1}{c}{}  
            & \multicolumn{1}{l}{{Word-CNN+MPAT}} 
            & \multicolumn{1}{c}{{\textbf{85.2}}} 
            & \multicolumn{1}{c}{\textbf{91.6}} 
            & \multicolumn{1}{c}{{\textbf{88.1}}} 
            & \multicolumn{1}{c}{{\textbf{6.72}}} 
            & \multicolumn{1}{c}{{\textbf{84.8}}} 
            & \multicolumn{1}{c}{{\textbf{21.2}}} 
            & \multicolumn{1}{c}{{\textbf{83.3}}} 
            & \multicolumn{1}{c}{{\textbf{25.2}}} \\
        %\cline{3-10}
        \multicolumn{1}{c}{}  
            & \multicolumn{1}{l}{{BERT}} 
            & \multicolumn{1}{c}{{91.9}} 
            & \multicolumn{1}{c}{{91.7}} 
            & \multicolumn{1}{c}{{59.4}} 
            & \multicolumn{1}{c}{{44.6}} 
            & \multicolumn{1}{c}{{43.7}} 
            & \multicolumn{1}{c}{{56.3}} 
            & \multicolumn{1}{c}{{11.2}} 
            & \multicolumn{1}{c}{{87.4}} \\
        \multicolumn{1}{c}{}  
            & \multicolumn{1}{l}{{BERT+MPAT}} 
            & \multicolumn{1}{c}{{\textbf{92.3}}} 
            & \multicolumn{1}{c}{{\textbf{92.3}}} 
            & \multicolumn{1}{c}{{\textbf{84.9}}} 
            & \multicolumn{1}{c}{{\textbf{11.7}}} 
            & \multicolumn{1}{c}{{\textbf{86.0}}} 
            & \multicolumn{1}{c}{{\textbf{6.8}}} 
            & \multicolumn{1}{c}{{\textbf{79.6}}} 
            & \multicolumn{1}{c}{{\textbf{24.5}}} \\
        \hline
        
        \multicolumn{1}{l}{\multirow{8}{*}{{AGNEWS}}}  
            & \multicolumn{1}{l}{{LSTM}} 
            & \multicolumn{1}{c}{{89.0}} 
            & \multicolumn{1}{c}{{90.4}} 
            & \multicolumn{1}{c}{{60.2}} 
            & \multicolumn{1}{c}{{34.3}} 
            & \multicolumn{1}{c}{{79.9}} 
            & \multicolumn{1}{c}{{19.7}} 
            & \multicolumn{1}{c}{{63.8}} 
            & \multicolumn{1}{c}{{28.4}} \\
        \multicolumn{1}{l}{}  
            & \multicolumn{1}{l}{{LSTM+MPAT}} 
            & \multicolumn{1}{c}{{\textbf{90.5}}} 
            & \multicolumn{1}{c}{{\textbf{92.6}}} 
            & \multicolumn{1}{c}{{\textbf{85.2}}} 
            & \multicolumn{1}{c}{{\textbf{8.39}}} 
            & \multicolumn{1}{c}{{\textbf{89.3}}} 
            & \multicolumn{1}{c}{{\textbf{10.9}}} 
            & \multicolumn{1}{c}{{\textbf{86.1}}} 
            & \multicolumn{1}{c}{{\textbf{11.6}}} \\
        \multicolumn{1}{l}{}  
            & \multicolumn{1}{l}{{Bi-LSTM}} 
            & \multicolumn{1}{c}{{90.5}} 
            & \multicolumn{1}{c}{{91.3}} 
            & \multicolumn{1}{c}{{61.9}} 
            & \multicolumn{1}{c}{{35.6}} 
            & \multicolumn{1}{c}{{79.9}} 
            & \multicolumn{1}{c}{{18.8}} 
            & \multicolumn{1}{c}{{66.0}} 
            & \multicolumn{1}{c}{{27.7}} \\
        \multicolumn{1}{l}{}  
            & \multicolumn{1}{l}{{Bi-LSTM+MPAT}} 
            & \multicolumn{1}{c}{{\textbf{91.6}}} 
            & \multicolumn{1}{c}{{\textbf{93.4}}} 
            & \multicolumn{1}{c}{{\textbf{88.3}}} 
            & \multicolumn{1}{c}{{\textbf{8.09}}} 
            & \multicolumn{1}{c}{{\textbf{91.2}}} 
            & \multicolumn{1}{c}{{\textbf{9.4}}} 
            & \multicolumn{1}{c}{{\textbf{91.4}}} 
            & \multicolumn{1}{c}{{\textbf{12.2}}} \\
        \multicolumn{1}{l}{}  
            & \multicolumn{1}{l}{{Word-CNN}} 
            & \multicolumn{1}{c}{{89.9}} 
            & \multicolumn{1}{c}{{94.0}} 
            & \multicolumn{1}{c}{{59.6}} 
            & \multicolumn{1}{c}{{37.3}} 
            & \multicolumn{1}{c}{{83.4}} 
            & \multicolumn{1}{c}{{15.6}} 
            & \multicolumn{1}{c}{{63.3}} 
            & \multicolumn{1}{c}{{32.7}} \\
        \multicolumn{1}{l}{}  
            & \multicolumn{1}{l}{{Word-CNN+MPAT}} 
            & \multicolumn{1}{c}{{\textbf{91.8}}} 
            & \multicolumn{1}{c}{{\textbf{94.8}}} 
            & \multicolumn{1}{c}{{\textbf{89.3}}} 
            & \multicolumn{1}{c}{{\textbf{6.87}}} 
            & \multicolumn{1}{c}{{\textbf{92.5}}} 
            & \multicolumn{1}{c}{{\textbf{8.3}}} 
            & \multicolumn{1}{c}{{\textbf{93.1}}} 
            & \multicolumn{1}{c}{{\textbf{11.5}}} \\
        \multicolumn{1}{l}{}  
            & \multicolumn{1}{l}{{BERT}} 
            & \multicolumn{1}{c}{{91.6}} 
            & \multicolumn{1}{c}{{91.0}} 
            & \multicolumn{1}{c}{{71.8}} 
            & \multicolumn{1}{c}{{33.9}} 
            & \multicolumn{1}{c}{{62.6}} 
            & \multicolumn{1}{c}{{37.7}} 
            & \multicolumn{1}{c}{{25.8}} 
            & \multicolumn{1}{c}{{75.2}} \\
        \multicolumn{1}{l}{}  
            & \multicolumn{1}{l}{{BERT+MPAT}} 
            & \multicolumn{1}{c}{{\textbf{91.8}}} 
            & \multicolumn{1}{c}{{\textbf{91.3}}} 
            & \multicolumn{1}{c}{{\textbf{89.2}}} 
            & \multicolumn{1}{c}{{\textbf{12.9}}} 
            & \multicolumn{1}{c}{{\textbf{88.1}}} 
            & \multicolumn{1}{c}{{\textbf{4.8}}} 
            & \multicolumn{1}{c}{{\textbf{81.5}}} 
            & \multicolumn{1}{c}{{\textbf{20.3}}} \\
        \hline
        
        \multicolumn{1}{l}{\multirow{4}{*}{{SNLI}}}  
            & \multicolumn{1}{l}{{ESIM}} 
            & \multicolumn{1}{c}{{87.2}} 
            & \multicolumn{1}{c}{{\textbf{88.2}}} 
            & \multicolumn{1}{c}{{69.4}} 
            & \multicolumn{1}{c}{{24.1}} 
            & \multicolumn{1}{c}{{56.2}} 
            & \multicolumn{1}{c}{{41.5}} 
            & \multicolumn{1}{c}{{21.6}} 
            & \multicolumn{1}{c}{{80.1}} \\
        \multicolumn{1}{l}{}  
            & \multicolumn{1}{l}{{ESIM+MPAT}} 
            & \multicolumn{1}{c}{{\textbf{87.4}}} 
            & \multicolumn{1}{c}{{{87.5}}}
            & \multicolumn{1}{c}{{\textbf{84.3}}} 
            & \multicolumn{1}{c}{{\textbf{15.1}}} 
            & \multicolumn{1}{c}{{\textbf{81.6}}} 
            & \multicolumn{1}{c}{{\textbf{17.7}}} 
            & \multicolumn{1}{c}{{\textbf{83.7}}} 
            & \multicolumn{1}{c}{{\textbf{15.6}}} \\
        \multicolumn{1}{l}{}  
            & \multicolumn{1}{l}{{BERT}} 
            & \multicolumn{1}{c}{{89.1}} 
            & \multicolumn{1}{c}{{89.6}} 
            & \multicolumn{1}{c}{{71.3}} 
            & \multicolumn{1}{c}{{29.2}} 
            & \multicolumn{1}{c}{{58.5}} 
            & \multicolumn{1}{c}{{44.6}} 
            & \multicolumn{1}{c}{{24.7}} 
            & \multicolumn{1}{c}{{74.0}} \\
        \multicolumn{1}{l}{}  
            & \multicolumn{1}{l}{{BERT+MPAT}} 
            & \multicolumn{1}{c}{{\textbf{91.7}}} 
            & \multicolumn{1}{c}{{\textbf{89.8}}} 
            & \multicolumn{1}{c}{{\textbf{87.7}}} 
            & \multicolumn{1}{c}{{\textbf{10.2}}} 
            & \multicolumn{1}{c}{{\textbf{82.7}}} 
            & \multicolumn{1}{c}{{\textbf{13.2}}} 
            & \multicolumn{1}{c}{{\textbf{78.3}}} 
            & \multicolumn{1}{c}{{\textbf{25.4}}} \\
        \specialrule{.1em}{0em}{.1em}
    \end{tabular}
% \vspace{-0.2cm}
\label{tab:defense performance}

\end{table*}

\subsubsection{Attack Methods} 
We choose three of the most representative black-box attacks to attack the above victim model. As the three methods can implement malicious adversarial attacks and are demonstrated to be more effective than or comparable to many other methods.
%--------------对比实验结果-----------------%
\begin{table*}[t]
    \centering
    \caption{\label{tab:compare}
    BPAT VS. SAFER VS. MPAT for the model accuracy on BERT and the attack success rate under PWWS, TextFooler and BertAttack. The best performance are shown in \textbf{bold}. Note that since the model accuracy here is all on the adversarial examples, ACC$_{adv}$ is referred to as ACC for short.}
\begin{tabular}{llcccccc}
    \specialrule{.1em}{0em}{.1em}
    \multicolumn{1}{l}{\textbf{Dataset}}
        &\multicolumn{1}{l}{\textbf{Model}}
        &\multicolumn{2}{c}{\textbf{PWWS(\%)}}
        &\multicolumn{2}{c}{\textbf{TextFooler(\%)}}
        &\multicolumn{2}{c}{\textbf{BertAttack(\%)}}\\
    \multicolumn{1}{l}{}
        &\multicolumn{1}{l}{}
        & \textbf{ACC$\uparrow$} 
        & \textbf{ASR$\downarrow$} 
        & \textbf{ACC$\uparrow$} 
        & \textbf{ASR$\downarrow$} 
        & \textbf{ACC$\uparrow$} 
        & \textbf{ASR$\downarrow$}  \\ 
    \hline
    \multicolumn{1}{l}{~}
        &Vanilla &59.4 &44.6 &43.7 &56.3 &11.2 &87.4\\
    \multicolumn{1}{l}{~}
        &BPAT    &69.3 &31.8 &84.1 &9.6 &34.7 &77.4\\
    \multicolumn{1}{l}{IMDB}
        &SAFER   &74.2 &28.6 &84.9 &11.8 &26.5 &88.5\\
    \multicolumn{1}{l}{~}
        &MPAT    &\textbf{84.9} &\textbf{11.7}
                 &\textbf{86.0} &\textbf{6.8}
                 &\textbf{79.6} &\textbf{24.5}\\
    \hline
    \multicolumn{1}{l}{~}
        &Vanilla &71.8 &33.9 &62.6 &37.7 &25.8 &75.2\\
    \multicolumn{1}{l}{~}
        &BPAT    &83.1 &26.9 &85.3 &6.2 &61.7 &39.5\\
    \multicolumn{1}{l}{AGNEWS}
        &SAFER   &88.9 &24.8 &\textbf{90.1} &5.6 &40.1 &67.9\\
    \multicolumn{1}{l}{~}
        &MPAT    &\textbf{89.2} &\textbf{12.9}
                 &88.1&\textbf{4.8}
                 &\textbf{81.5} &\textbf{20.3}\\
    \hline
    \multicolumn{1}{l}{~}
        &Vanilla &71.3 &29.3 &58.5 &44.6 &24.7 &74.0\\
    \multicolumn{1}{l}{~}
        &BPAT    &61.4 &31.1 &68.5 &33.5 &22.7 &74.5\\
    \multicolumn{1}{l}{SNLI}
        &SAFER   &74.1 &18.8 &72.8 &23.1 &69.1 &29.6\\
    \multicolumn{1}{l}{~}
        &MPAT    &\textbf{87.7} &\textbf{10.2}
                 &\textbf{82.7} &\textbf{13.2}
                 &\textbf{78.3} &\textbf{25.4}\\
    \specialrule{.1em}{0em}{.1em}
    \end{tabular}
\end{table*}
\begin{itemize}
      \item 
      \textit{PWWS}: \cite{PWWS} performs synonym replacements or substitutes named entities with other similar entities. Since PWWS has no named entity replacement rules specially designed for NLI tasks, we only applied synonym replacements on SNLI without named entities substitutions.
      \item  
      \textit{TextFooler}: \cite{TextFooler} utilizes counter-fitting word embeddings from \cite{Counter} for synonym replacements. 
      \item  
       \textit{BertAttack}: applies BERT in a semantic-preserving manner to generate substitutes for words \cite{BertAttack}. By utilizing the contextual understanding provided by BERT, the generated adversarial examples maintain the overall semantic coherence of the original input.
\end{itemize}
\subsubsection{Baseline Models} 
We compare our MPAT with three baseline methods.
\begin{itemize}
      \item 
      \textit{Vanilla}: Normal training without any defense methods is performed to establish a baseline. By using the vanilla model, we can assess the accuracy and compare it with the performance of models employing defense methods on the original test data.
      \item  
      \textit{BPAT}: Adversarial training with benign perturbations, we use the same training configuration as FREEAT\cite{FREEAT}, and dynamically add benign perturbations during the training process. 
      \item  
       \textit{SAFER}: \cite{SAFER} proposes a randomized smoothing based method that constructs a stochastic ensemble by applying random word substitutions on the input sentences.
\end{itemize}
\subsubsection{Hyper-parameters} 
We determine the values of $r$ and $\epsilon$ through a series of empirical experiments, resulting in $r=35\%$ and $\epsilon= 0.0005$. The experimental results can be found in Section \ref{sec:hyper}. In the experiments, we set the value of $\lambda$ to 1. Additionally, to save computational resources, we choose an ascent step of $K=3$.
\subsection{Defense Performance} 
We evaluate the performance of different defense methods using the following metrics.
\begin{itemize}
    \item \textbf{ACC}: Model accuracy, which is the ratio between the number of correctly predicted examples and the total number of testing examples;
    \item \textbf{ASR}: Attack success rate, the ratio of the number of adversarial examples which cause incorrect predictions to the total number of adversarial examples.
\end{itemize}
The higher ACC and lower ASR after being attacked, the better the defense performance.
\par In order to evaluate the performance, we perform a uniform sampling of examples from each class in the original test set, resulting in the creation of 1000 clean examples for each dataset. Subsequently, diverse attack methods are employed to generate corresponding adversarial examples based on these clean examples. Finally, ACC and ASR are measured under different settings, as presented in Table \ref{tab:defense performance} and \ref{tab:compare}.
\subsubsection{Our MPAT}
Experimental results in Table \ref{tab:defense performance} demonstrate that MPAT is capable of building a robust model, ensuring the following: 
\begin{itemize}
    \item 
    Preservation of performance on the original task: When there is no attack, the accuracy ACC$_{test}$ on test examples are mostly improved after applying our defense method. As shown in Table \ref{tab:defense performance}, except for the setting of SNLI dataset and ESIM victim model, the ACC$_{test}$ under other settings shows an improvement ranging from 0.1 to 3.5. This improvement proves that our MPAT effectively maintains a high level of accuracy on the original test set, demonstrating its ability to provide robust defense without compromising the model's performance on the original task; 
    \item
    Effective resistance against adversarial attacks: When there is an attack, no matter what the experimental settings are, after our MPAT, not only the model accuracy is noticeably improved, but the attack success rate is also significantly reduced. This result demonstrates the capability of MPAT in effectively resisting and mitigating the impact of adversarial attacks. As an example, when LSTM is targeted by BertAttack, the accuracy ACC$_{adv}$ on the IMDB dataset substantially increases, rising from less than 26\% to over 90\%. 
\end{itemize}
%In addition, when the victim models are attacked under adversarial attack, our MPAT has significant defense performance.
%Experimental results show that our defense method can effectively reduce the attack success rate on both IMDB and AGNEWS datasets. 
%We also compare BPAT and SAFER with our MPAT as shown in Table \ref{table:vs}. 
%Please note that we evaluate the defense performance only on BERT because: (1) Bert is the most widely used DNN model; and (2) the experimental results in [] show that SAFER has the higher certified accuracy on BERT than Word-CNN. It can be seen from the third to fourth columns of Table 4 that: (1) For the model’s accuracy after attack, our methods exceed BPAT and SAFER by about 10\% and 15\% respectively on IMDB; and (3) For the attack success rate, our method has reached the minimum value. To this end, our experiments suggest that MPAT is superior to other AT methods at both ACC and ASR and works well to build a more robust model.
\subsubsection{MPAT VS. SAFER VS. BPAT}
We also compare BPAT and SAFER with our MPAT as shown in Table \ref{tab:compare}.
We specifically evaluate the defense performance on the BERT model for the following reasons: (1) BERT is one of the most widely used pre-trained language models. Many studies and application scenarios employ BERT as a benchmark model for performance evaluation and comparison; and (2) the experimental results in \cite{SAFER} show that SAFER has the higher certified accuracy on BERT than Word-CNN.

It can be seen from Table \ref{tab:compare} that MPAT achieves the overall best performance among all methods: (1) For the model accuracy after attack, our methods exceed BPAT and SAFER by about 2\% to 55\% and 1\% to 53\% respectively on different settings; and (2) For the attack success rate, our method achieves the minimum value under all settings. 
Therefore, we can draw the conclusion that  MPAT shows superior capability to protect models from adversarial attacks and achieves comparable top performance on the three datasets, where it leads to further improvement in ACC and significant reductions in ASR.

\subsubsection{Tests of statistical significance}
\label{sec:significance}
\begin{figure*}[t]
  \centering
  \begin{minipage}[b]{0.3\textwidth}
    \centering
    \label{fig:sig1}
    \includegraphics[width=\textwidth]{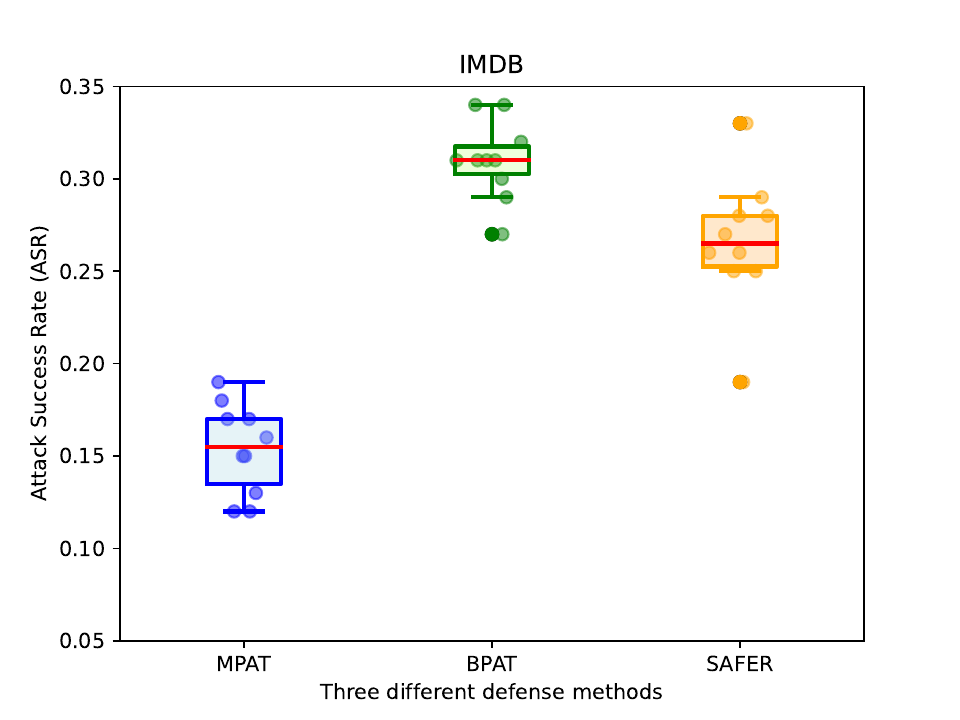}
    \text{(a)}
  \end{minipage}
  \begin{minipage}[b]{0.3\textwidth}
    \centering
    %\label{fig:sig2}
    \includegraphics[width=\textwidth]{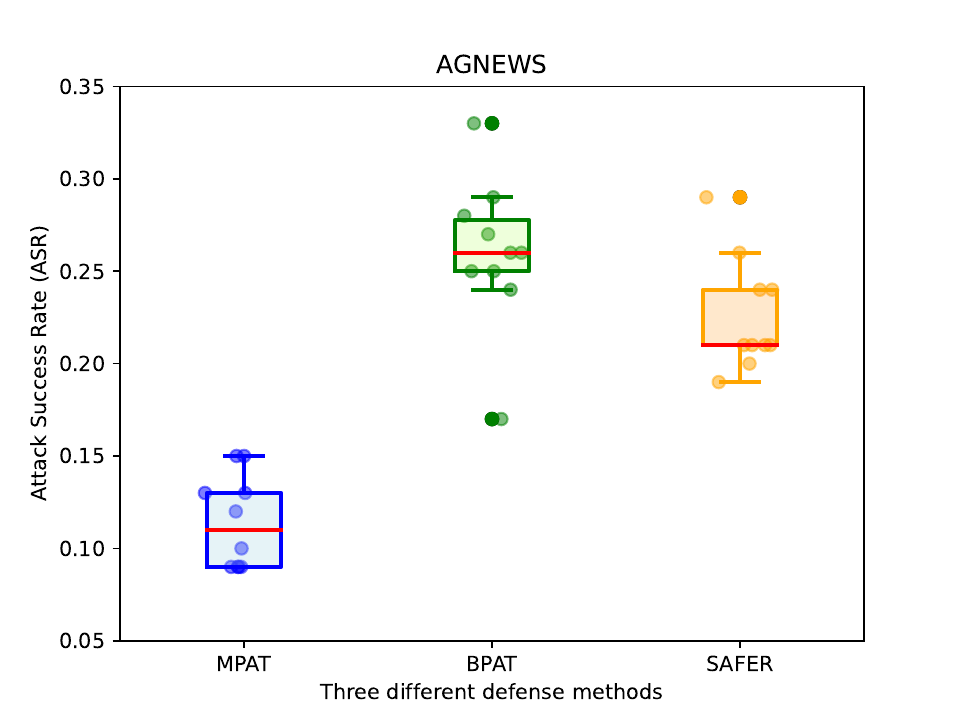}
    \text{(b)}
  \end{minipage}
  \begin{minipage}[b]{0.3\textwidth}
    \centering
    \includegraphics[width=\textwidth]{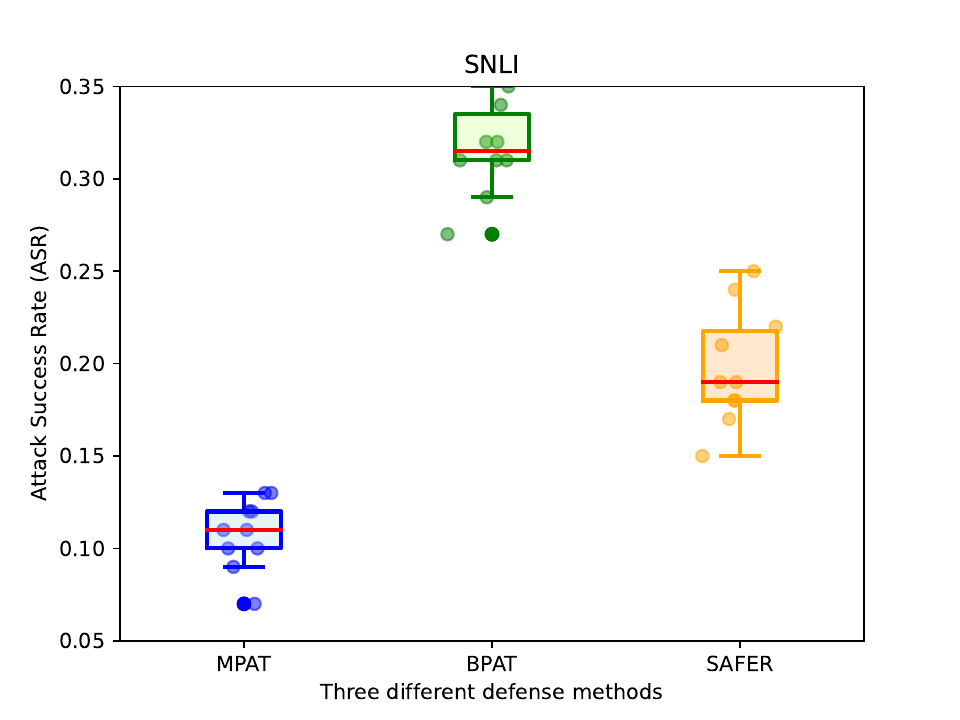}
    \text{(c)}
    %\label{fig:sig3}
  \end{minipage}
  \caption{Distribution of ASRs corresponding to each baseline. (a), (b) and (c) show the distribution of examples under the IMDB, AGNEWS and SNLI respectively, where the lower the distribution of data points, the better the baseline defense performance.}
  \label{fig:sig}
\end{figure*}

%---------画表格展示t、h------------%
\begin{table}[t]
    \centering
    \caption{\label{tab:sig}
    Experimental results of independent t-test. The larger the t-value and the smaller the p-value, the more confident we can be in asserting the significant difference between the two baselines.}
    \begin{tabular}{llcc}
    \specialrule{.1em}{0em}{.1em}
    \multicolumn{1}{l}{\textbf{Dataset}}
        &\multicolumn{1}{l}{\textbf{Baselines}}
        &\multicolumn{1}{c}{\textbf{T-value}}
        &\multicolumn{1}{c}{\textbf{P-value}}\\
    \hline
    \multicolumn{1}{l}{IMDB}
        &\multicolumn{1}{l}{{MPAT VS. SAFER}}
        &\multicolumn{1}{c}{{8.36}}
        &\multicolumn{1}{c}{{1.56$e$-05}}\\
    \multicolumn{1}{l}{}
        &\multicolumn{1}{l}{{MPAT VS. BPAT}}
        &\multicolumn{1}{c}{{12.32}}
        &\multicolumn{1}{c}{{6.17$e$-07}}\\
    \hline
    \multicolumn{1}{l}{AGNEWS}
        &\multicolumn{1}{l}{{MPAT VS. SAFER}}
        &\multicolumn{1}{c}{{12.39}}
        &\multicolumn{1}{c}{{5.88$e$-07}}\\
    \multicolumn{1}{l}{}
        &\multicolumn{1}{l}{{MPAT VS. BPAT}}
        &\multicolumn{1}{c}{{7.89}}
        &\multicolumn{1}{c}{{2.48$e$-05}}\\
    \hline
    \multicolumn{1}{l}{{SNLI}}
        &\multicolumn{1}{l}{{MPAT VS. SAFER}}
        &\multicolumn{1}{c}{{8.62}}
        &\multicolumn{1}{c}{{1.21$e$-05}}\\
    \multicolumn{1}{l}{}
        &\multicolumn{1}{l}{{MPAT VS. BPAT}}
        &\multicolumn{1}{c}{{25.63}}
        &\multicolumn{1}{c}{{1.01$e$-09}}\\
    \specialrule{.1em}{0em}{.1em}
    \end{tabular}
\end{table}
We first conduct 10 non-repeated samplings from each datasets, resulting in a total of 3*10 clean datasets, each consisting of 1000 examples. For each dataset, we apply PWWS to attack BERT and obtain ten sets of ASR results. 
Based on these results, an \textbf{Independent T-Test} is performed to determine the significance of the differences in defense performance between MPAT and the other two baseline methods (Table \ref{tab:sig}). And corresponding boxplots are drawn to show the distribution and statistical indicators of any two independent sets of ASR values under different settings (Fig. \ref{fig:sig}).
%The significant T-values and extremely small P-values on all three datasets in Table \ref{tab:sig} show that: (1) MPAT demonstrates a notable distinction from BPAT, and (2) MPAT exhibits a significant contrast from SAFER.
%Hence, it can be inferred that there are genuine differences between the two baselines, and these differences are not due to randomness.
\par The experimental results in Table \ref{tab:sig} indicate that for all datasets: (1) the calculated t-values exceed the critical values ($\pm1.96$), and (2) the corresponding p-values are smaller than the predetermined significance level ($0.05$). Therefore, the significant T-values and extremely small P-values on all three datasets in Table \ref{tab:sig} show that there are significant differences in defense performance between MPAT and BPAT, as well as between MPAT and SAFER.

Meanwhile, the results depicted in Fig. \ref{fig:sig} further provide visual evidence of significant differences: By examining the distribution patterns and central tendencies of the examples, it is evident that our MPAT method consistently exhibits the lowest ASR values on all datasets. Furthermore, these values are significantly lower compared to the other two baseline methods.
%---------消融实验IMDB/AGNEWS---------%
\begin{table}[t]
    \centering
   \caption{\label{tab:Ablation:IMDB/AGNEWS}
    Ablation experiment for the accuracy (ACC) and the attack success rate (ASR) on IMDB and AGNEWS.
    }
    \begin{tabular}{lcccc}
    \specialrule{.1em}{0em}{.1em}
    \textbf{Victim Model}
        &\multicolumn{2}{c}{\textbf{IMDB(\%)}}
        &\multicolumn{2}{c}{\textbf{AGNEWS(\%)}}\\
    \multicolumn{1}{c}{~}
        &\textbf{ACC$\uparrow$}
        &\textbf{ASR$\downarrow$}
        &\textbf{ACC$\uparrow$}
        &\textbf{ASR$\downarrow$}\\
    \hline
    LSTM + $MPAT$&\textbf{83.9}&\textbf{10.0}&\textbf{85.2}&\textbf{8.39}\\
    LSTM + $Only~Para.$&44.5&55.5&61.0&33.4\\
    LSTM + $Only~Syn.$&80.3&12.4&81.5&10.3\\
    LSTM + $Only~BPAT$&23.2&76.6&64.9&30.4\\
    LSTM + $Syn.~and~BPAT$&82.3&10.3&83.9&8.41\\
    \hline
    Bi-LSTM + $MPAT$&\textbf{86.9}&\textbf{8.83}&86.7&8.09\\
    Bi-LSTM + $Only~Para.$&49.6&50.4&67.9&32.2\\
    Bi-LSTM + $Only~Syn.$&81.1&11.7&84.5&10.2\\
    Bi-LSTM + $Only~BPAT$&29.5&69.9&69.8&26.9\\
    Bi-LSTM + $Syn.~and~BPAT$&85.8&9.91&\textbf{88.3}&\textbf{6.91}\\
    \hline
    Word-CNN + $MPAT$&\textbf{88.0}&\textbf{0.10}&\textbf{90.4}&\textbf{6.87}\\
    Word-CNN + $Only~Para.$&38.0&53.5&61.8&36.6\\
    Word-CNN + $Only~Syn.$&78.8&12.6&86.4&10.15\\
    Word-CNN + $Only~BPAT$&20.7&75.1&64.9&23.6\\
    Word-CNN + $Syn.~and~BPAT$&85.4&12.9&89.3&6.99\\
    \hline
    BERT + $MPAT$&\textbf{83.9}&\textbf{17.0}&\textbf{89.2}&12.9\\
    BERT + $Only~Para.$&60.4&40.3&73.3&19.9\\
    BERT + $Only~Syn.$&72.1&18.6&85.0&11.5\\
    BERT + $Only~BPAT$&70.4&51.8&84.1&24.8\\
    BERT + $Syn.~and~BPAT$&78.6&18.2&83.6&\textbf{10.2}\\
    \specialrule{.1em}{0em}{.1em}
    \end{tabular}
\end{table}
%---------显著差异实验结果-------------%
%---------消融实验SNLI---------%
\begin{table}[t]
    \centering
    \caption{\label{tab:Ablation:SNLI}
    Ablation experiment for the accuracy (ACC) and the attack success rate (ASR) on SNLI.
    }
    \begin{tabular}{lcc}
    \specialrule{.1em}{0em}{.1em}
    \textbf{Victim Model}
        &\textbf{ACC}
        &\textbf{ASR}\\
    \hline
    ESIM + $MPAT$&\textbf{84.3}&\textbf{15.1}\\
    ESIM + $Only~Para.$&83.5&17.6\\
    ESIM + $Only~Syn.$&79.3&15.8\\
    ESIM + $Only~BPAT$&70.4&23.3\\
    ESIM + $Syn.~and~BPAT$&77.7&15.8\\
    \hline
    BERT + $MPAT$&\textbf{87.7}&\textbf{10.2}\\
    BERT + $Only~Para.$&75.7&17.1\\
    BERT + $Only~Syn.$&77.6&16.2\\
    BERT + $Only~BPAT$&61.4&31.1\\
    BERT + $Syn.~and~BPAT$&82.9&11.9\\
    \specialrule{.1em}{0em}{.1em}
    \end{tabular}
\end{table}
\subsection{Ablation Study}
We conduct ablation experiments on each component of MPAT to investigate its effectiveness. The results for IMDB and AGNEWS are summarized in Table \ref{tab:Ablation:IMDB/AGNEWS} and the results for SNLI are summarized in Table \ref{tab:Ablation:SNLI}, where (1) $only~Para.$ refers to the defense which only applies text paraphrase during model training; (2) $only~Syn.$ refers to the defense which only applies synonym replacement during model training; (3) $only~BPAT$ refers to the benign adversarial training method named FreeAT; (4) $Syn.~and~BPAT$ refers to our method without text paraphrase; and (5) $MPAT$ refers to our proposed MPAT method. 
\par As shown in Table \ref{tab:Ablation:IMDB/AGNEWS}, for the IMDB dataset, MPAT demonstrates superior defense performance compared to the scenarios where various components of MPAT are removed, with improvements observed in both ACC and ASR metrics. For the AGNEWS dataset, MPAT performs well in terms of the ACC metric, while the ASR metric shows a slight decrease in the Bi-LSTM and BERT models. Table \ref{tab:Ablation:SNLI} illustrates that MPAT achieves excellent performance in both metrics for the SNLI dataset.
%The experimental results in Tables \ref{tab:Ablation:IMDB/AGNEWS} and \ref{tab:Ablation:SNLI} show that the defense performance will decrease when any component of MPAT is removed. 
We discuss the experimental results in the following two points.
\subsubsection{The respective contributions of the three perturbations}
We first study the importance of the three perturbations when they are applied individually. 
Among the three level perturbations, perturbation at the word-level (synonym replacement) plays a key role in enhancing the model's resistance, and perturbation at the sentence-level (text paraphrase) follows closely and contribute significantly to the robustness of the model, while the performance of perturbation at the embedding-level (BPAT) is relatively weak. Part of the reasons for this phenomenon are: 1) By incorporating synonym replacement, the model becomes more adaptive to malicious adversarial perturbation, thereby improving its defensive performance; 2) The attack model is based on word replacement, so synonym perturbation is more effective than paraphrase perturbation; 3) The purpose of BPAT is to enhance the robustness of the model to benign natural perturbations, and it is difficult to effectively resist maliciously crafted adversarial examples. 
\subsubsection{The necessity of text paraphrase}
To explore the role of text paraphrase in enhancing model robustness, we compare the accuracy and the attack success rate of MPAT with and without paraphrase perturbation. The results show that in most cases, text paraphrase can further improve the defense effect after adding benign adversarial perturbation and synonym replacement perturbation. By introducing paraphrase perturbation, the model becomes more capable of adapting to variations in adversarial examples, including changes in sentence structure, semantic expressions, and more. Paraphrase perturbation allows the model to better handle different forms of adversarial examples, ensuring that it remains robust against a wide range of attacks. Thus combining all three perturbations, MPAT achieves the best performance.
\begin{figure}[t]
    \centering
    \begin{minipage}[b]{0.4\textwidth}
        \centering
        \includegraphics[width=\textwidth]{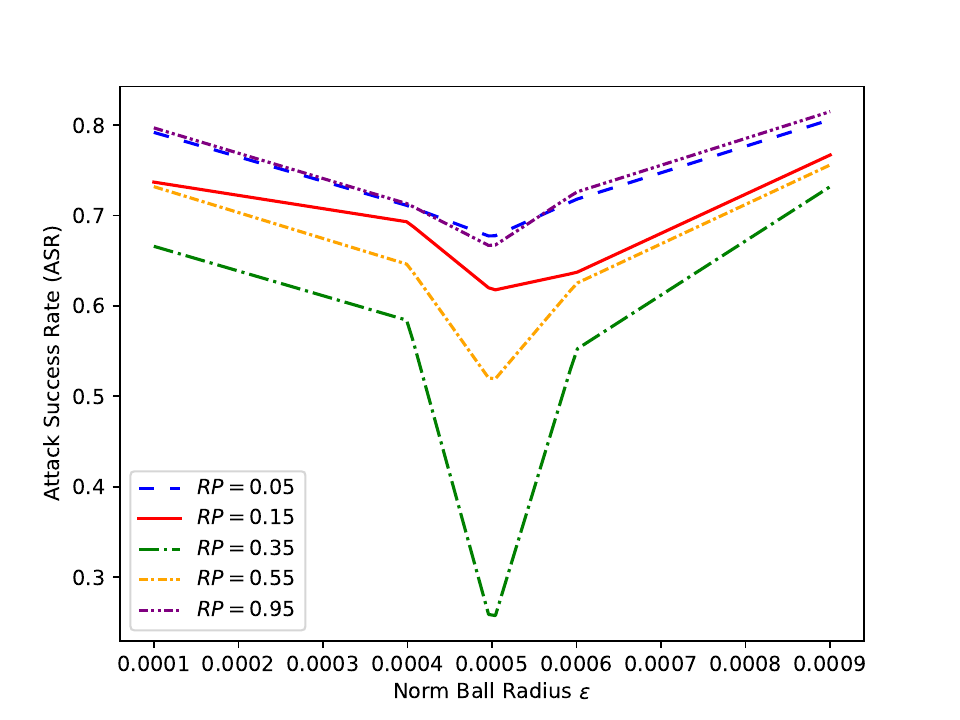}
        \text{(a)}
    \end{minipage}
     \begin{minipage}[b]{0.4\textwidth}
        \centering
        \includegraphics[width=\textwidth]{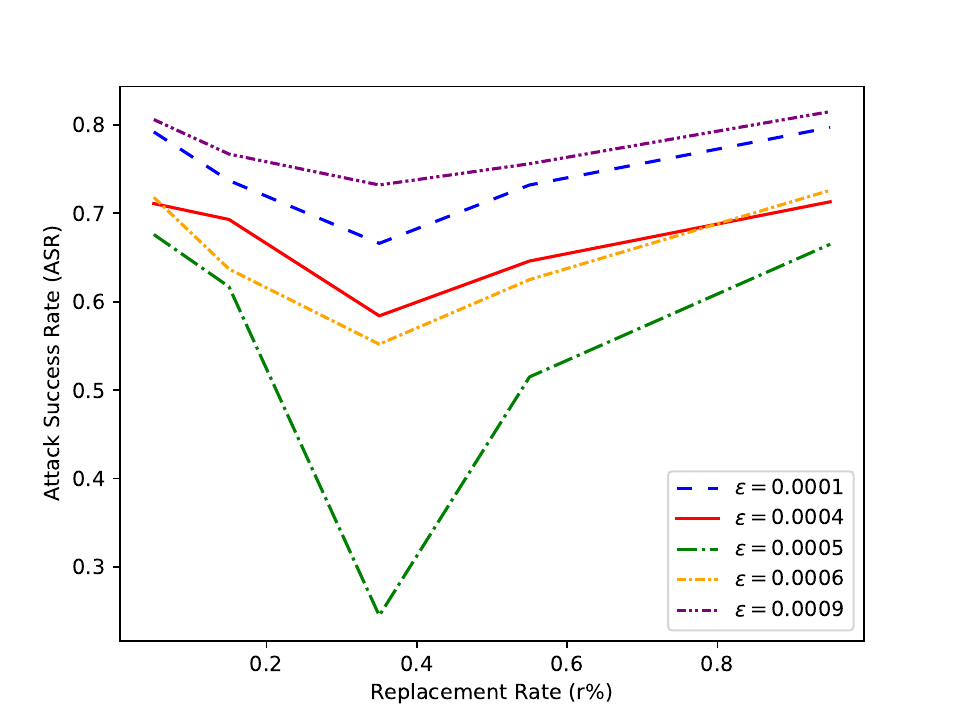}  
        \text{(b)}
  \end{minipage}
   \caption{Comparison of defense performance under different replacement rates $r\%$ and $\epsilon$-ball radii $\epsilon$.}
    \label{fig:rreps}
\end{figure}
\subsection{Hyper-parameter Tuning}
\label{sec:hyper}
%--------rp和epsilon--------%
\begin{table*}[t]
 \centering
 \caption{Adversarial examples where our method successfully defends, but the baseline method SAFER fails to defend. Clean means the clean examples, and Adv. is the abbreviation for adversarial example. Note that the underlined words are modified by the attacker, $srr$ refers to the synonym replacement rate.}
 \label{tab:result_adv_example}
 \tabcolsep 4pt
 {\scriptsize
   \begin{tabular}{p{2.2cm}p{13cm}p{1.5cm}}
    \specialrule{.1em}{0em}{.1em}
        \textbf{Type}&\multicolumn{1}{l}{\textbf{Example}}&\textbf{Prediction}  \\ 
        \hline
        Clean
        & 18 dead after Japanese earthquakes. The death toll from a series of powerful earthquakes in central Japan has reached 18, with more than 800 injured. The biggest quake measured 6.8 on the Richter scale.
        &  World \\  
        \hline
        Adv. against SAFER 
        & 18 \textcolor{myred}{\underline{stagnant}} after Japanese earthquakes. The \textcolor{myred}{\underline{end}} \textcolor{myred}{\underline{price}} from a series of \textcolor{myred}{\underline{hefty}} earthquakes in central Japan has reached 18, with more than 800 \textcolor{myred}{\underline{wound}}. The biggest \textcolor{myred}{\underline{temblor}} measured 6.8 on the Richter \textcolor{myred}{\underline{scurf}} . \textbf{[Defense failed with $srr=0.184$]}
         &  \colorbox{red}{Business} \\ \hline 
         Adv. against MPAT
        & 18 \textcolor{mygreen}{\underline{bushed}} after Japanese earthquakes. The \textcolor{mygreen}{\underline{end}} \textcolor{mygreen}{\underline{cost}} from a series of \textcolor{mygreen}{\underline{hefty}} earthquakes in \textcolor{mygreen}{\underline{key}} \textcolor{mygreen}{\underline{Nihon}} has \textcolor{mygreen}{\underline{hit}} 18, with more than 800 \textcolor{mygreen}{\underline{hurt}}. The biggest \textcolor{mygreen}{\underline{temblor}} measured 6.8 on the Richter \textcolor{mygreen}{\underline{exfoliation}}. \textbf{[Successful defense with $srr=0.263$]}
         &  \colorbox{green}{World} \\ \hline\hline

      Clean
        & Bush Predicted No Iraq Casualties, Robertson Says. The Rev. Pat Robertson said President Bush dismissed his warning that the United States would suffer heavy casualties in Iraq and told the television evangelist just before the beginning of the war that we're not going to have any casualties.
        &  World \\  
        \hline
        Adv. against SAFER 
        & \textcolor{myred}{\underline{Shrub}} Predicted No Iraq Casualties, Robertson Says. The \textcolor{myred}{\underline{rpm.}} \textcolor{myred}{\underline{dab}} Robertson said \textcolor{myred}{\underline{prexy}} \textcolor{myred}{\underline{shrub}} dismissed his \textcolor{myred}{\underline{admonition}} that the United States would suffer \textcolor{myred}{\underline{grievous}} casualties in \textcolor{myred}{\underline{Irak}} and told the television evangelist just before the \textcolor{myred}{\underline{rootage}} of the war that we 're not going to \textcolor{myred}{\underline{ingest}} any casualties. \textbf{[Defense failed with $srr=0.185$]}
         &  \colorbox{red}{Sports} \\ \hline 
         Adv. against MPAT
        & \textcolor{mygreen}{\underline{Dubya}} Predicted No Iraq Casualties, \textcolor{mygreen}{\underline{Oscar}} Says. The \textcolor{mygreen}{\underline{rpm.}} \textcolor{mygreen}{\underline{dab}} \textcolor{mygreen}{\underline{Oscar}} said President \textcolor{mygreen}{\underline{shrub}} dismissed his \textcolor{mygreen}{\underline{monition}} that the United \textcolor{mygreen}{\underline{DoS}} would \textcolor{mygreen}{\underline{meet}} \textcolor{mygreen}{\underline{large}} casualties in \textcolor{mygreen}{\underline{Irak}} and told the \textcolor{mygreen}{\underline{video}} evangelist \textcolor{mygreen}{\underline{barely}} before the \textcolor{mygreen}{\underline{root}} of the \textcolor{mygreen}{\underline{state}} that we 're not going to \textcolor{mygreen}{\underline{throw}} any casualties. \textbf{[Successful defense with $srr=0.327$]}
         &  \colorbox{green}{World} \\ \specialrule{.1em}{0em}{.1em}

   \end{tabular}
  }
\end{table*}

In this experiment, we set the perturbation $\epsilon$-ball radius $\epsilon\in[0.0001,0.001)$ and the word replacement rate $r\in (0,1)$. We conducted 25 non-repetitive experiments using 25 different combinations within these ranges. Each experiment was based on a specific set of hyper-parameters, ensuring that each combination was thoroughly evaluated and compared. With this experimental design, we were able to comprehensively explore the effects of different perturbation radius and replacement rates on defense performance and find the optimal hyper-parameter configuration.

Fig. \ref{fig:rreps}(a) shows the impact of different perturbation ball radius on ASR, with each line representing a different replacement rate. From the figure, it can be observed that the green curve corresponding to a specific replacement rate achieves the lowest attack success rate. This indicates that, under the same perturbation ball radius, selecting an appropriate replacement rate can effectively reduce the likelihood of successful attacks. Fig. \ref{fig:rreps}(b), on the other hand, displays the influence of different replacement rates on ASR, with each line representing a different perturbation ball radius. Similarly, the green curve corresponding to a specific perturbation ball radius achieves the lowest attack success rate. This suggests that, under the same replacement rate, choosing an appropriate perturbation ball radius can provide better defense performance. Taking the results from both Fig. \ref{fig:rreps}(a) and Fig. \ref{fig:rreps}(b) into consideration, we can conclude that selecting a replacement rate of $35\%$ and a perturbation ball radius of $0.0005$ yields the lowest attack success rate, thus achieving optimal defense performance.

\section{Discussion}
\label{sec:discussions}

\subsection{Case study}
Table \ref{tab:result_adv_example} presents the adversarial examples generated using PWWS attack on the AGNEWS dataset. We applied our method MPAT and the baseline method SAFER to generated adversarial examples from the same clean examples. When attacking BERT that has been reinforced by our defense method, attack method PWWS based on synonym replacement cannot successfully attack even after putting in more effort, which requires more synonym replacement rate. However, for the model after reinforcement using the baseline method, attackers only need to use a lower synonym replacement rate to successfully attack.
\par We have also discovered that adversarial examples, which can render our defense methods ineffective, frequently include hate speech. Such instances often involve language that evokes strong emotions, provocative statements, or misleading information, as shown in Table \ref{tab:failexample}. 
\begin{table}[t]
 \centering
 \caption{Adversarial examples where our method fails to defend. Note that \underline{\textbf{w$*$ore$*$on$*$er}} is an offensive word that we replaced manually with asterisks.}
 \label{tab:failexample}
   \begin{tabular}{lp{5.8cm}}
    \specialrule{.1em}{0em}{.1em}
        \textbf{Type} & \textbf{Example}\\ 
        \hline
        Clean&Key debate ahead in US election. US President George W Bush and Democratic challenger John Kerry are preparing for a key TV debate.\\
        \hline
    Adv. against MPAT&Key debate ahead in \textcolor{myred}{\underline{uracil}} election. \textcolor{myred}{\underline{uracil}} \textcolor{myred}{\underline{prexy}} George W \textcolor{myred}{\underline{Vannevar}} and Democratic challenger \textcolor{myred}{\underline{\textbf{w$*$ore$*$on$*$er}}} kerry \textcolor{myred}{\underline{exist}} preparing for a key TV \textcolor{myred}{\underline{argumentation}}. ~~\textbf{[Defense failed]}  \\   
        \specialrule{.1em}{0em}{.1em}
   \end{tabular}
\end{table}
Attackers may intentionally choose these emotionally conflicting and controversial texts as attack examples to increase the effectiveness of adversarial examples. 
%We conclude that the defense failure is due to the significant semantic difference between the adversarial example and the original text, leading to their divergence from the same distribution.
We have explored the reasons why our MPAT fails to defend against these adversarial examples, as explained below. An input example $x$ which is fed into DNNs is generated by a natural physical process with a certain probability, rather than being artificially created or fabricated. For example, in text classification tasks such as news classification, the process of generating training data rarely produces hate speech. However, the classifier does not have the option to reject such an example and is forced to output a class label. Therefore, one of the reasons why adversarial examples can disrupt the classifier is because they are far from the boundary of the manifold of the task.
\subsection{Future Work}
Given that the current mainstream adversarial attack methods are mostly based on synonym replacement, in order to further explore the ability of paraphrasing perturbation to improve the robustness of the model to deal with malicious adversarial attacks, our future research and improvement directions can focus on the following two aspects: First, in-depth Study the mechanism of sentence-level attacks, analyze the attacker's attack strategy and perturbation methods on the model, in order to better understand the nature and characteristics of sentence-level attacks. Secondly, we can further improve the model's robustness to sentence-level adversarial attacks, and enhance the model's resistance to malicious attacks by introducing more defense mechanisms and perturbation methods.

\section{Conclusion}
We propose MPAT, a malicious perturbation based adversarial training method, to build robust DNN model against malicious adversarial attacks in NLP. Specifically, we construct a multi-level malicious example generation strategy using text paraphrase and synonym replacement, and implement adversarial training to minimize adversarial loss and additional manifold loss. Extensive experiments show that our method can effectively resist attacks while maintaining the original task performance compared with previous defense methods. In the future, we will further explore the potential of using malicious perturbation based adversarial training method to improve the robustness of the model and resist more kinds of adversarial attacks.

\bibliographystyle{IEEEtran}
\bibliography{MPAT.bib}

\end{document}